\documentclass[10pt,twocolumn,letterpaper]{article}

\usepackage{iccv}
\usepackage{times}
\usepackage{epsfig}
\usepackage{graphicx}
\usepackage{amsmath}
\usepackage{amssymb}

\usepackage{epstopdf}
\usepackage{amsmath}
\usepackage{amssymb}
\usepackage{appendix}
\usepackage{balance}
\usepackage{amsfonts}
\usepackage[linesnumbered, ruled]{algorithm2e}
\SetKwRepeat{Do}{do}{while}%
\usepackage{url}
\usepackage{setspace}
\usepackage[english]{babel}
\usepackage{multirow}
\usepackage{color}
\usepackage[bb=boondox,cal=boondoxo]{mathalfa}
\usepackage[export]{adjustbox}
\usepackage{subcaption}
\usepackage{enumitem}
\usepackage{color}
\usepackage{float}
\usepackage{caption}
\usepackage{stfloats}

\usepackage[pagebackref=true,breaklinks=true,letterpaper=true,colorlinks,bookmarks=false]{hyperref}

\iccvfinalcopy 

\def\iccvPaperID{2643} 

\ificcvfinal\fi

\begin{document}

\title{Deep Clustering via Joint Convolutional Autoencoder Embedding and Relative Entropy Minimization}

\author{Kamran Ghasedi Dizaji${}^\dag$, Amirhossein Herandi${}^\ddag$, Cheng Deng${}^\sharp$, Weidong Cai${}^\natural$, Heng Huang${}^\dag$\thanks{Corresponding Author. This work (accepted at ICCV 2017) was partially supported by U.S. NIH R01 AG049371, NSF IIS 1302675, IIS 1344152, DBI 1356628,
IIS 1619308, IIS 1633753.}\\
${}^\dag$Electrical and Computer Engineering, University of Pittsburgh, USA\\
${}^\ddag$Computer Science and Engineering, University of Texas at Arlington, USA\\
${}^\sharp$School of Electronic Engineering, Xidian University, China\\
${}^\natural$School of Information Technologies, University of Sydney, Australia\\
{\tt\small kamran.ghasedi@gmail.com, amirhossein.herandi@uta.edu, chdeng@mail.xidian.edu.cn}\\
{\tt\small tom.cai@sydney.edu.au, heng.huang@pitt.edu}
}

\maketitle

\begin{abstract}
	\vspace{-0.2cm}
	Image clustering is one of the most important computer vision applications, which has been extensively studied in literature. However, current clustering methods mostly suffer from lack of efficiency and scalability when dealing with large-scale and high-dimensional data.
	In this paper, we propose a new clustering model, called \textbf{DE}e\textbf{P} Embedded Regular\textbf{I}zed \textbf{C}lus\textbf{T}ering (\textit{DEPICT}), which efficiently maps data into a discriminative embedding subspace and precisely predicts cluster assignments. \textit{DEPICT} generally consists of a multinomial logistic regression function stacked on top of a multi-layer convolutional autoencoder. We define a clustering objective function using relative entropy (\textit{KL} divergence) minimization, regularized by a prior for the frequency of cluster assignments. An alternating strategy is then derived to optimize the objective by updating parameters and estimating cluster assignments. Furthermore, we employ the reconstruction loss functions in our autoencoder, as a data-dependent regularization term, to prevent the deep embedding function from overfitting. In order to benefit from end-to-end optimization and eliminate the necessity for layer-wise pretraining,  we introduce a joint learning framework to minimize the unified clustering and reconstruction loss functions together and train all network layers simultaneously. Experimental results indicate the superiority and faster running time of \textit{DEPICT} in real-world clustering tasks, where no labeled data is available for hyper-parameter tuning.
\end{abstract}

\vspace{-0.5cm}
\section{Introduction}
\vspace{-0.1cm}
\begin{figure}[!t]
	\begin{subfigure}{.16\textwidth}
		\centering
		\includegraphics[width=\linewidth]{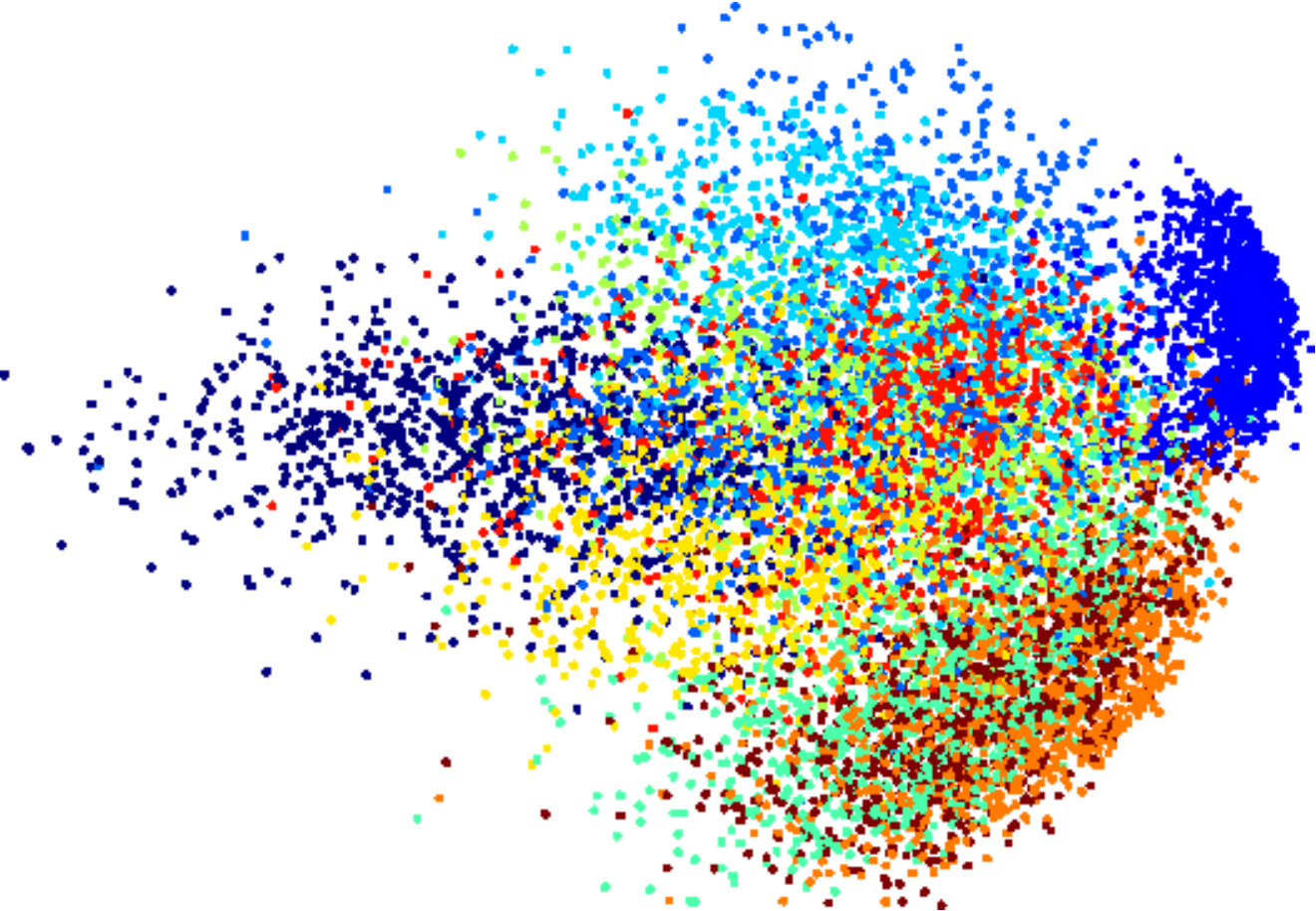}
		\caption{Raw Data}
	\end{subfigure}%
	\begin{subfigure}{.16\textwidth}
		\centering
		\includegraphics[width=\linewidth]{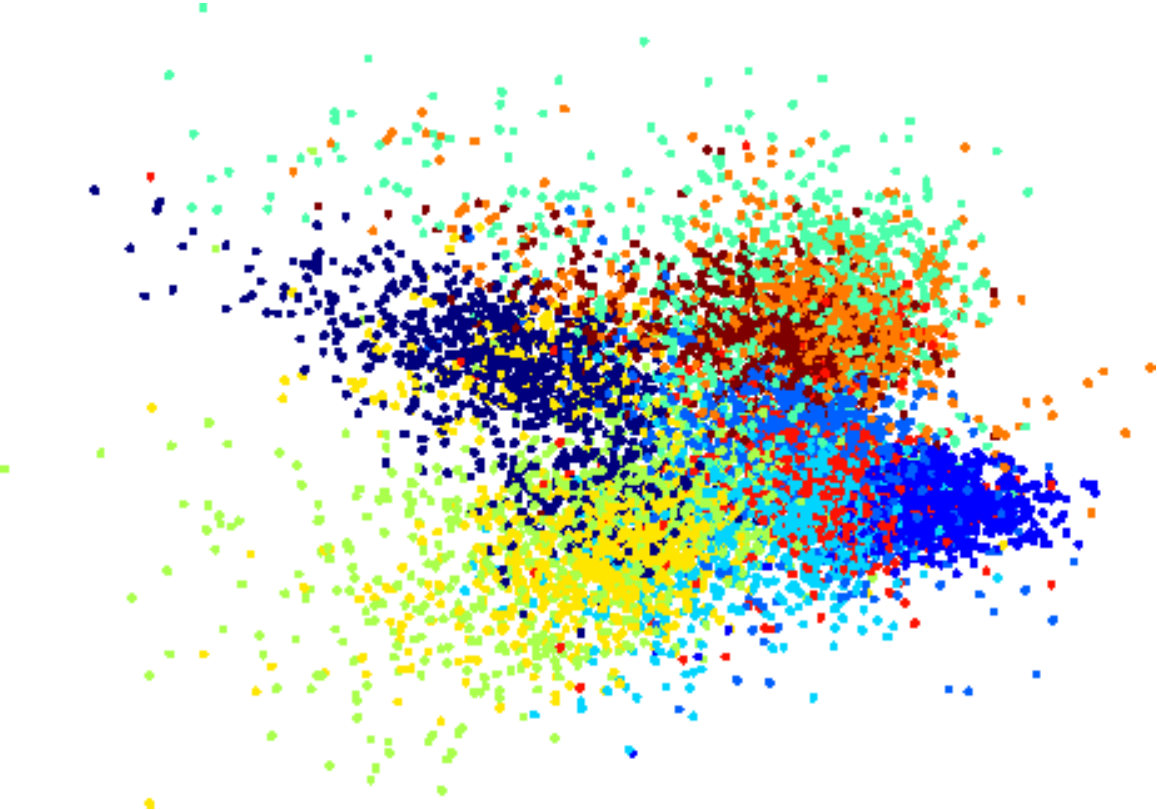}
		\caption{NonJoint \textit{DEPICT}}
	\end{subfigure}
	\begin{subfigure}{.153\textwidth}
		\centering
		\includegraphics[trim=40mm 30mm 30mm 30mm, clip,  width=1\linewidth]{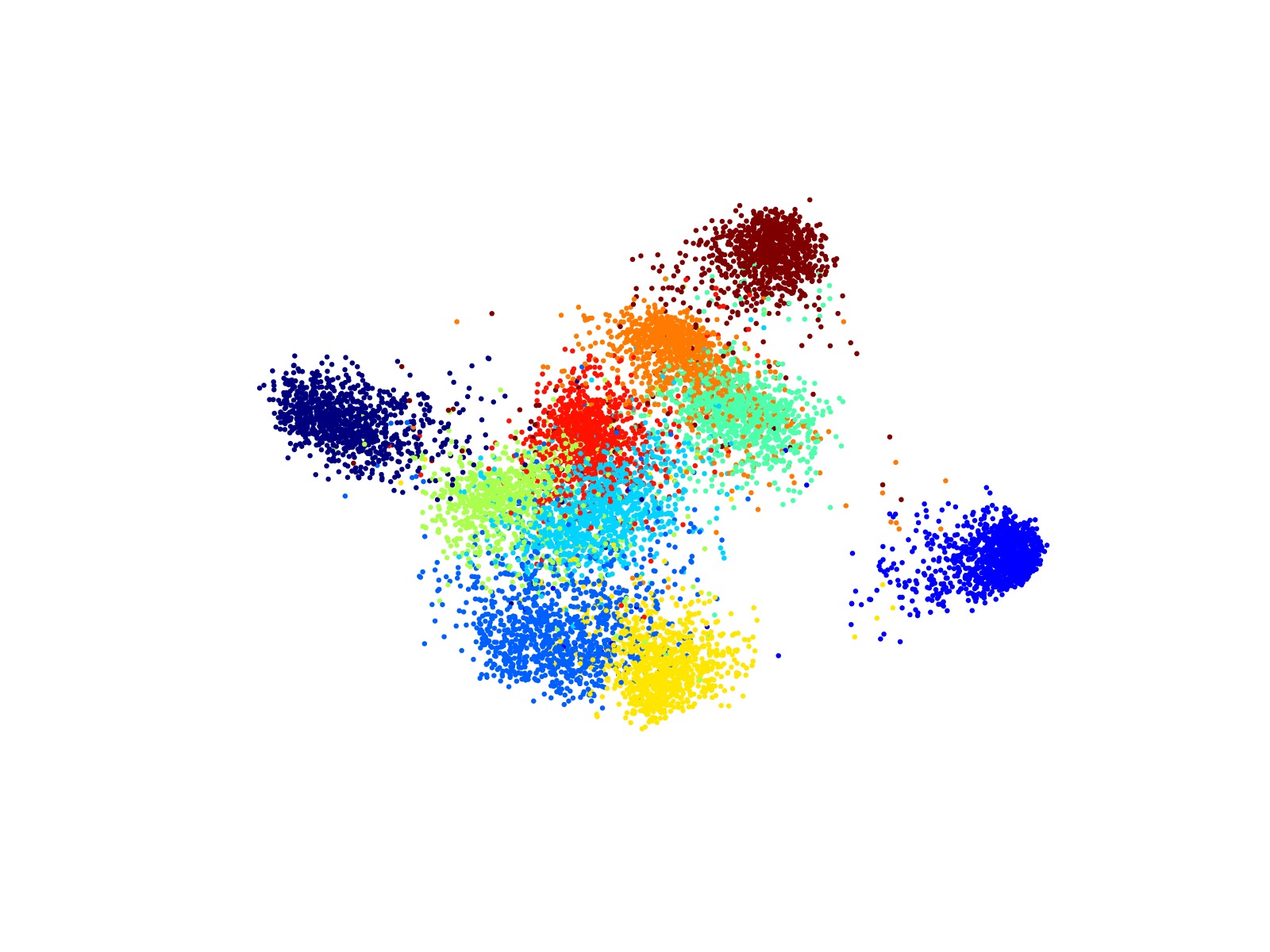}
		\caption{Joint \textit{DEPICT}}
	\end{subfigure}
	\caption{Visualization to show the discriminative capability of embedding subspaces using \textit{MNIST-test} data. (a) The space of raw data.  (b) The embedding subspace of non-joint \textit{DEPICT} using standard stacked denoising autoencoder (SdA). (c) The embedding subspace of joint \textit{DEPICT} using our joint learning approach (MdA).}
	\label{fig_1}
\end{figure}

Clustering is one of the fundamental research topics in machine learning and computer vision research, and it has gained significant attention for discriminative representation of data points without any need for supervisory signals. The clustering problem has been extensively studied in various applications; however, the performance of standard clustering algorithms is adversely affected when dealing with high-dimensional data, and their time complexity dramatically increases when working with large-scale datasets. Tackling the curse of dimensionality, previous studies often initially project data into a low-dimensional manifold, and then cluster the embedded data in this new subspace  \cite{roth2003feature,tian2014learning,wang2016learning}. Handling large-scale datasets, there are also several studies which select only a subset of data points to accelerate the clustering process \cite{shinnou2008spectral,Dijun:ICDE10,Huang2015AAAI2}.

However, dealing with real-world image data, existing clustering algorithms suffer from different issues: 1) Using inflexible hand-crafted features, which do not depend on the input data distribution; 2) Using shallow and linear embedding functions, which are not able to capture the non-linear nature of data; 3) Non-joint embedding and clustering processes, which do not result in an optimal embedding subspace for clustering; 4) Complicated clustering algorithms that require tuning the hyper-parameters using labeled data, which is not feasible in real-world clustering tasks.

To address the mentioned challenging issues, we propose a new clustering algorithm, called deep embedded regularized  clustering (\textit{DEPICT}), which exploits the advantages of both discriminative clustering methods and deep embedding models. \textit{DEPICT} generally consists of two  main parts, a multinomial logistic regression (soft-max) layer stacked on top of a multi-layer convolutional autoencoder. The soft-max layer along with the encoder pathway can be considered as a discriminative clustering model, which is trained using the relative entropy (\textit{KL} divergence) minimization. We further add a regularization term based on a prior distribution for the frequency of cluster assignments. The regularization term penalizes unbalanced cluster assignments and prevents allocating clusters to outlier samples.

Although this deep clustering model is flexible enough to discriminate the complex real-world input data,  it can easily get stuck in non-optimal local minima during training and result in undesirable cluster assignments. In order to avoid overfitting the deep clustering model to spurious data correlations, we utilize the reconstruction loss function of autoencoder models as a data-dependent regularization term for training parameters.

In order to benefit from a joint learning framework for embedding and clustering, we introduce a unified objective function including our clustering and auxiliary reconstruction loss functions. We then employ an alternating approach to efficiently update the parameters and estimate the cluster assignments. It is worth mentioning that in the standard learning approach for training a multi-layer autoencoder, the encoder and decoder parameters are first pretrained layer-wise using the reconstruction loss, and the encoder parameters are then fine-tuned using the objective function of the main task \cite{vincent2010stacked}. However, it has been argued that the non-joint fine-tuning step may overwrite the encoder parameters entirely and consequently cancel out the benefit of the layer-wise pretraining step \cite{zhao2015stacked}. To avoid this problem and achieve optimal joint learning results, we simultaneously train all of the encoder and decoder layers together along with the soft-max layer. To do so, we sum up the squared error reconstruction loss functions between the decoder and their corresponding (clean) encoder layers and add them to the clustering loss function.

Figure \ref{fig_1} demonstrates the importance of our joint learning strategy by comparing different data representations of \textit{MNIST-test} data points \cite{lecun1998gradient} using principle component analysis (PCA) visualization. The first figure indicates the raw data representation; The second one shows the data points in  the embedding subspace of non-joint \textit{DEPICT}, in which the model is trained using the standard layer-wise stacked denoising autoencoder (SdA); The third one visualizes the data points in the embedding subspace of joint \textit{DEPICT}, in which the model is trained using our multi-layer denoising autoencoder learning approach (MdA). As shown, joint \textit{DEPICT} using MdA learning approach provides a significantly more discriminative embedding subspace compared to non-joint \textit{DEPICT} using standard SdA learning approach.

Moreover, experimental results show that \textit{DEPICT} achieves superior or competitive results compared to the state-of-the-art algorithms on the image benchmark datasets while having faster running times. In addition, we compared different learning strategies for \textit{DEPICT}, and confirm that our joint learning approach has the best results. It should also be noted that  \textit{DEPICT} does not require any hyper-parameter tuning using supervisory signals, and consequently is a better candidate for the real-world clustering tasks.
Thus, we summarize the advantages of \textit{DEPICT} as:
\begin{itemize}[noitemsep,topsep=2pt]
	\item Providing a discriminative non-linear embedding subspace via the deep convolutional autoencoder;
	\item Introducing an end-to-end joint learning approach, which  unifies the clustering and embedding tasks, and avoids layer-wise pretraining;
	\item Achieving superior or competitive clustering results on high-dimensional and large-scale datasets with no need for hyper-parameter tuning using labeled data.
\end{itemize}

\section{Related Works}\label{section:realted_works}
There is a large number of clustering algorithms in literature, which can be grouped into different perspectives, such as hierarchical \cite{heller2005bayesian,williams1999mcmc,zhang2012graph}, centroid-based \cite{lloyd1982least,bezdek1984fcm,Huang2014KDD2,bahmani2012scalable}, graph-based \cite{shi2000normalized,nie2016constrained,wang2016structured,Huang2016AAAI5}, sequential (temporal) \cite{keogh2001online,sargin2008analysis,sadoughi2015msp,zhou2013hierarchical,sadoughi2015retrieving}, regression model based \cite{Huang2016ICDM1,Huang2013ICML1}, and subspace clustering models \cite{agrawal1998automatic,kailing2004density,Huang2015ICCV,Huang2016IJCAI1}. In another sense, they are generally divided into two subcategories, generative and discriminative clustering algorithms. The generative algorithms like \textit{$K$-means} and Gaussian mixture model \cite{biernacki2000assessing} explicitly represent the clusters using geometric properties of the feature space, and model the categories via the statistical distributions of input data. Unlike the generative clustering algorithms, the discriminative methods directly identify the categories using their separating hyperplanes regardless of data distribution. Information theoretic \cite{li2004minimum,barber2005kernelized,krause2010discriminative}, max-margin \cite{zhao2008efficient,xu2004maximum}, and spectral graph \cite{ng2002spectral} algorithms are examples of discriminative clustering models. Generally it has been argued that the discriminative models often have better results compared to their generative counterparts, since they have fewer assumptions about the data distribution and directly separate the clusters, but their training can suffer from overfitting or getting stuck in undesirable local minima \cite{krause2010discriminative,ng2002spectral,raina2003classification}. Our \textit{DEPICT} algorithm is also a discriminative clustering model, but it benefits from the auxiliary reconstruction task of autoencoder to alleviate this issues in training of our discriminative clustering algorithm.

There are also several studies regarding the combination of clustering with feature embedding learning. Ye \emph{et al.} introduced a kernelized \textit{$K$-means} algorithm, denoted by \textit{DisKmeans}, where embedding to a lower dimensional subspace via linear discriminant analysis (\textit{LDA}) is jointly learned with \textit{$K$-means} cluster assignments \cite{ye2008discriminative}. \cite{Huang2014ECML1} proposed to a new method to simultaneously conduct both clustering and feature embedding/selection tasks to achieve better performance. But these models suffer from having shallow and linear embedding functions, which cannot represent the non-linearity of real-world data.

A joint learning framework for updating code books and estimating image clusters was proposed in \cite{xie2015integrating} while \textit{SIFT} features are used as input data. A deep structure, named \textit{TAGnet} was introduced in \cite{wang2016learning}, where two layers of sparse coding followed by a clustering algorithm are trained with an alternating learning approach. Similar work is presented in \cite{wang2015joint} that formulates a joint optimization framework for discriminative clustering and feature extraction using sparse coding. However, the inference complexity of sparse coding forces the model in \cite{wang2015joint} to reduce the dimension of input data with \textit{PCA} and the model in \cite{wang2016learning} to use an approximate solution. Hand-crafted features and dimension reduction techniques degrade the clustering performance by neglecting the distribution of input data.

Tian \emph{et al.} learned a non-linear embedding of the affinity graph using a stacked autoencoder, and then obtained the clusters in the embedding subspace via \textit{$K$-means} \cite{tian2014learning}.
Trigeorgis \emph{et al.} extended semi non-negative matrix factorization (\textit{semi-NMF}) to stacked multi-layer (deep) \textit{semi-NMF} to capture the abstract information in the top layer.  Afterwards, they run \textit{$K$-means} over the embedding subspace for cluster assignments \cite{trigeorgis2014deep}.
More recently, Xie \emph{et al.} employed denoising stacked autoencoder learning approach, and first pretrained the model layer-wise and then fine-tuned the encoder pathway stacked by a clustering algorithm using Kullback-Leibler divergence minimization \cite{xie2016unsupervised}. Unlike these models that require  layer-wise pretraining as well as non-joint embedding and clustering learning, \textit{DEPICT} utilizes an end-to-end optimization for training all network layers simultaneously using the unified clustering and reconstruction loss functions.

Yang \emph{et al.} introduced a new clustering model, named \textit{JULE}, based on a recurrent framework, where data is represented via a convolutional neural network and embedded data is iteratively clustered using an agglomerative clustering algorithm \cite{yang2016joint}. They derived a unified loss function consisting of the merging process for agglomerative clustering and updating the parameters of the deep representation. While \textit{JULE} achieved good results using the joint learning approach, it requires tuning of a large number of hyper-parameters, which is not practical in real-world clustering tasks. In contrast, our model does not need any supervisory signals for hyper-parameter tuning.

\section{Deep Embedded Regularized Clustering} \label{section:method}
In this section, we first introduce the clustering objective function and the corresponding optimization algorithm, which alternates between estimating the cluster assignments and updating model parameters. Afterwards, we show the architecture of \textit{DEPICT} and provide the joint learning framework to simultaneously train all network layers using the unified clustering and reconstruction loss functions.


\subsection{DEPICT Algorithm} \label{section:RLC}

Let's consider the clustering task of $N$ samples, $\mathbf{X} = [\mathbf{x}_1, ..., \mathbf{x}_n]$, into $K$ categories, where each sample $\mathbf{x}_i \in \mathbb{R}^{d_x}$.  Using the embedding function, $\varphi_{W}: X\rightarrow Z$, we are able to map raw samples into the embedding subspace $\mathbf{Z} = [\mathbf{z}_1, ..., \mathbf{z}_n]$, where each $\mathbf{z}_i \in \mathbb{R}^{d_z}$ has a much lower dimension compared to the input data (i.e. $d_z \ll d_x$). Given the embedded features, we use a multinomial logistic regression (soft-max) function $f_{\theta}: Z\rightarrow Y$ to predict the probabilistic cluster assignments as follows.
\begin{align} \label{eq:1}
&	p_{ik} = P(y_i=k | \mathbf{z}_i, \boldsymbol{\Theta})= \frac{exp( \boldsymbol{\theta}_k^T \mathbf{z}_i )}{\sum\limits_{k'=1}^K exp ( \boldsymbol{\theta}_{k'}^T \mathbf{z}_i )} \,,
\end{align}
where $\boldsymbol{\Theta} = [\boldsymbol{\theta}_1, ..., \boldsymbol{\theta}_k] \in \mathbb{R}^{d_z \times  K}$ are the soft-max function parameters, and $p_{ik}$ indicates the probability of the $i$-th sample belonging to the $k$-th cluster.

In order to define our clustering objective function, we employ an auxiliary target variable $\mathbf{Q}$ to refine the model predictions iteratively. To do so, we first use Kullback-Leibler (\textit{KL}) divergence to decrease the distance between the model prediction $\mathbf{P}$ and the target variable $\mathbf{Q}$.
\begin{align} \label{eq:2}
\mathcal{L} = KL (\mathbf{Q} \| \mathbf{P}) = \frac{1}{N} \sum\limits_{i=1}^N \sum\limits_{k=1}^K q_{ik} \log  \frac{q_{ik}}{p_{ik}}  \,,
\end{align}
In order to avoid degenerate solutions, which allocate most of the samples to a few clusters or assign a cluster to outlier samples, we aim to impose a regularization term to the target variable. To this end, we first define the empirical label distribution of target variables as:
\begin{align} \label{eq:3}
&	f_{k} = P(\mathbf{y}=k) = \frac{1}{N} \sum\limits_{i} q_{ik}   \,,
\end{align}
where $f_k$ can be considered as the soft frequency of cluster assignments in the target distribution. Using this empirical distribution, we are able to enforce our preference for having balanced assignments by adding the following \textit{KL} divergence to the loss function.
\begin{align} \label{eq:4}
\mathcal{L} &= KL (\mathbf{Q} \| \mathbf{P})  + KL (\mathbf{f} \| \mathbf{u})  \\
&= \Big[ \frac{1}{N} \sum\limits_{i=1}^N \sum\limits_{k=1}^K q_{ik} \log  \frac{q_{ik}}{p_{ik}} \Big] +  \Big[ \frac{1}{N} \sum\limits_{k=1}^K f_{k} \log  \frac{f_{k}}{u_{k}} \Big] \nonumber \\
&= \frac{1}{N} \sum\limits_{i=1}^N \sum\limits_{k=1}^K q_{ik} \log  \frac{q_{ik}}{p_{ik}} +  q_{ik} \log  \frac{f_{k}}{u_{k}} \,, \nonumber
\end{align}
where $\mathbf{u}$ is the uniform prior for the empirical label distribution. While the first term in the objective minimizes the distance between the target and model prediction distributions, the second term balances the frequency of clusters in the target variables.
Utilizing the balanced target variables, we can force the model to have more balanced predictions (cluster assignments) $\mathbf{P}$ indirectly. It is also simple to change the prior from the uniform distribution to any arbitrary distribution in the objective function if there is any extra knowledge about the frequency of clusters.

An alternating learning approach is utilized to optimize the objective function. Using this approach,  we estimate the target variables $\mathbf{Q}$ via fixed parameters (expectation step), and update the parameters while the target variables $\mathbf{Q}$  are assumed to be known (maximization step). The problem to infer the target variable $\mathbf{Q}$ has the following objective:
\begin{align} \label{eq:5}
\min\limits_{\mathbf{Q}} \ \  \frac{1}{N} \sum\limits_{i=1}^N \sum\limits_{k=1}^K q_{ik} \log  \frac{q_{ik}}{p_{ik}} +  q_{ik} \log  \frac{f_{k}}{u_{k}} \,,
\end{align}
where the target variables are constrained to $\sum_k q_{ik} =1$. This problem can be solved using first order methods, such as gradient descent, projected gradient descent, and Nesterov optimal method \cite{nesterov2013introductory}, which only  require the objective function value and its (sub)gradient at each iteration. In the following equation, we show the partial derivative of the objective function with respect to the target variables.
\begin{align} \label{eq:6}
\frac{\partial  \mathcal{L}}{\partial q_{ik}} \propto  \log \Big( \frac{q_{ik} f_k}{p_{ik}} \Big) +  \frac{q_{ik}}{\sum\limits_{i'=1}^N q_{i'k}} + 1 \,,
\end{align}
Investigating this problem more carefully, we approximate the gradient in Eq.(\ref{eq:6}) by removing the second term, since the number of samples N is often big enough to ignore the second term. Setting the gradient equal to zero, we are now able to compute the closed form solution for $\mathbf{Q}$ accordingly.
\begin{align} \label{eq:7}
&	 q_{ik} =   \frac{  p_{ik}   / (\sum_{i'} p_{i'k})^{\frac{1}{2}}  }{\sum\limits_{k'}   p_{ik'}   / (\sum_{i'} p_{i'k'})^{\frac{1}{2}}}   \,,
\end{align}
For the maximization step, we update the network parameters $\boldsymbol{\psi} = \{\boldsymbol{\Theta}, \mathbf{W}\}$ using the estimated target variables with the following objective function.
\begin{align} \label{eq:8}
\min\limits_{\boldsymbol{\psi}} \ \  -\frac{1}{N} \sum\limits_{i=1}^N \sum\limits_{k=1}^K q_{ik} \log  p_{ik}\,,
\end{align}
Interestingly, this problem can be considered as a standard cross entropy loss function for classification tasks, and the parameters of soft-max layer $\boldsymbol{\Theta}$ and embedding function $\mathbf{W}$ can be efficiently updated by backpropagating the error.


\begin{figure*}[t]
	\begin{center}
		\includegraphics[trim=0mm 0mm 0mm 0mm, clip,  width=0.83\linewidth]{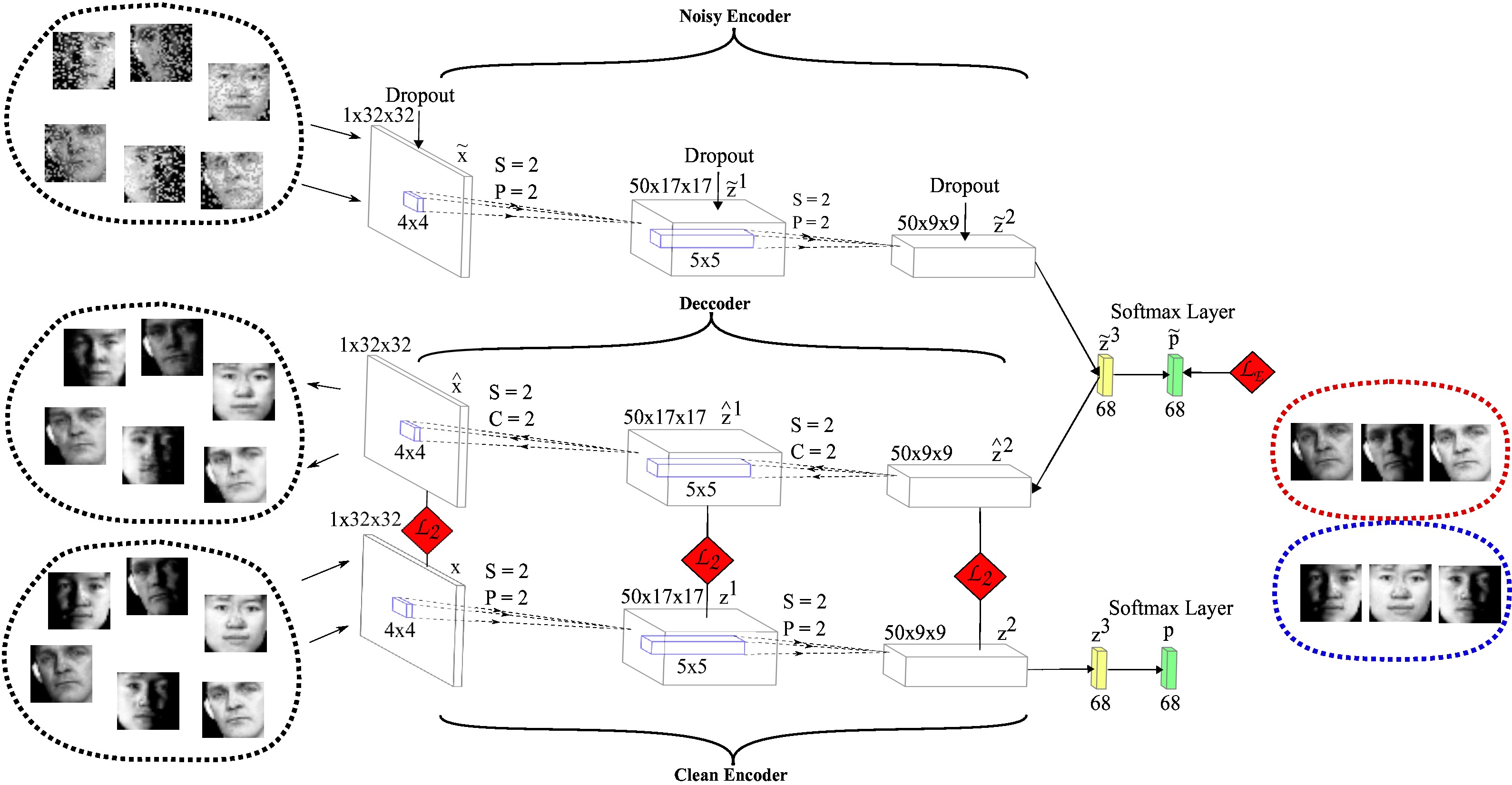}
	\end{center}
	\vspace{-0.3cm}
	\caption{Architecture of \textit{DEPICT} for \textit{CMU-PIE} dataset. \textit{DEPICT} consists of a soft-max layer stacked on top of a multi-layer convolutional autoencoder. In order to illustrate the joint learning framework, we consider the following four pathways for \textit{DEPICT}: Noisy (corrupted) encoder, Decoder, Clean encoder and Soft-max layer. The clustering loss function, $L_E$, is applied on the noisy pathway, and the reconstruction loss functions, $L_2$, are between the decoder and clean encoder layers. The output size of convolutional layers,  kernel sizes, strides (S), paddings (P) and crops (C) are also shown. }
	\vspace{-0.2cm}
	\label{fig:2}
\end{figure*}

\subsection{DEPICT Architecture}
In this section, we extend our general clustering loss function using a denoising autoencoder. The deep embedding function is useful for capturing the non-linear nature of input data; However, it may overfit to spurious data correlations and get stuck in undesirable local minima during training. To avoid this overfitting, we employ autoencoder structures and use the reconstruction loss function as a data-dependent regularization for training the parameters. Therefore, we design \textit{DEPICT} to consist of a soft-max layer stacked on top of a multi-layer convolutional autoencoder. Due to the promising performance of strided convolutional layers in \cite{radford2015unsupervised,yeh2016semantic}, we employ convolutional layers in our encoder and strided convolutional layers in the decoder pathways, and avoid deterministic spatial pooling layers (like max-pooling). Strided convolutional layers allow the network to learn its own spatial upsampling, providing a better generation capability.

Unlike the standard learning approach for denoising autoencoders, which contains layer-wise pretraining and then fine-tuning, we simultaneously learn all of the autoencoder and soft-max layers. As shown in Figure \ref{fig:2}, \textit{DEPICT} consists of the following components:

\noindent \textbf{1)} Corrupted feedforward (encoder) pathway maps the noisy input data into the embedding subspace using a few convolutional layers followed by a fully connected layer. The following equation indicates the output of each layer in the noisy encoder pathway.
\begin{align} \label{eq:50}
&	 \tilde{\mathbf{z}}^{l} = Dropout \big[g(\mathbf{W}_e^l \tilde{\mathbf{z}}^{l-1} ) \big] \,,
\end{align}
where $\tilde{\mathbf{z}}^{l}$ are the noisy features of the $l$-th layer, $Dropout$ is a stochastic mask function that randomly sets a subset of its inputs to zero \cite{srivastava2014dropout}, $g$ is the activation function of convolutional or fully connected layers, and $\mathbf{W}_e^l$ indicates the weights of the $l$-th layer in the encoder. Note that the first layer features, $\tilde{\mathbf{z}}^0$, are equal to the noisy input data, $\tilde{\mathbf{x}}$.

\noindent \textbf{2)} Followed by the corrupted encoder, the decoder pathway reconstructs the input data through a fully connected and multiple strided convolutional layers as follows,
\begin{align} \label{eq:51}
&	 \hat{\mathbf{z}}^{l-1} = g(\mathbf{W}_d^l \hat{\mathbf{z}}^{l} )  \,,
\end{align}
where $\hat{\mathbf{z}}^{l}$ is the $l$-th reconstruction layer output, and $\mathbf{W}_d^l$ shows the weights for the $l$-th layer of the decoder. Note that input reconstruction, $\hat{\mathbf{x}}$, is equal to $\hat{\mathbf{z}}^0$.

\noindent \textbf{3)} Clean feedforward (encoder) pathway shares its weights with the corrupted encoder, and infers the clean embedded features. The following equation shows the outputs of the clean encoder, which are used in the reconstruction loss functions and obtaining the final cluster assignments.
\begin{align} \label{eq:52}
&	 \mathbf{z}^{l} = g(\mathbf{W}_e^l \mathbf{z}^{l-1} )  \,,
\end{align}
where $\mathbf{z}^{l}$ is the clean output of the $l$-th layer in the encoder. Consider the first layer features $\mathbf{z}^0$ equal to input data $\mathbf{x}$.

\noindent \textbf{4)} Given the top layer of the corrupted and clean encoder pathways as the embedding subspace, the soft-max layer obtains the cluster assignments  using Eq.(\ref{eq:1}).

Note that we compute target variables $\mathbf{Q}$ using the clean pathway, and model prediction $\tilde{\mathbf{P}}$ via the corrupted pathway. Hence, the clustering loss function $KL (\mathbf{Q} \| \tilde{\mathbf{P}})$ forces the model to have invariant features with respect to noise. In other words, the model is assumed to have a dual role: a clean model, which is used to compute the more accurate target variables; and a noisy model, which is trained to achieve noise-invariant predictions.

As a crucial point, \textit{DEPICT} algorithm provides a joint learning framework that optimizes the soft-max and autoencoder parameters together.
\begin{align} \label{eq:53}
\min\limits_{\boldsymbol{\psi}} \ \  -\frac{1}{N} \sum\limits_{i=1}^N \sum\limits_{k=1}^K q_{ik} \log  \tilde{p}_{ik}  + \frac{1}{N}  \sum\limits_{i=1}^N  \sum\limits_{l=0}^{L-1} \frac{1}{|\mathbf{z}_i^l|} \| \mathbf{z}_i^l - \hat{\mathbf{z}}_i^l \|_2^2 \,,
\end{align}
where $|\mathbf{z}_i^l|$ is the output size of the $l$-th hidden layer (input for $l=0$), and $L$ is the depth of the autoencoder model.

The benefit of joint learning frameworks for training multi-layer autoencoders is also reported in semi-supervised  classification tasks \cite{rasmus2015semi,zhao2015stacked}. However, \textit{DEPICT} is different from previous studies, since it is designed for the unsupervised clustering task, it also does not require max-pooling switches used in stacked what-where autoencoder (SWWAE) \cite{zhao2015stacked}, and lateral (skip) connections between encoder and decoder layers used in ladder network \cite{rasmus2015semi}. Algorithm \ref{alg:1} shows a brief description of \textit{DEPICT} algorithm.

\begin{algorithm}[!t]
	Initialize $\mathbf{Q}$ using a clustering algorithm\\
	\While{not converged  }{
		$\min\limits_{\boldsymbol{\psi}} \ \  -\frac{1}{N} \sum\limits_{ik} q_{ik} \log  \tilde{p}_{ik}  + \frac{1}{N}  \sum\limits_{il} \frac{1}{|\mathbf{z}_i^l|} \| \mathbf{z}_i^l - \hat{\mathbf{z}}_i^l \|_2^2$\\
		$ p_{ik}^{(t)} \propto exp( \boldsymbol{\theta}_k^T \mathbf{z}_i^{L} )$  \\ 
		$ q_{ik}^{(t)}   \propto   p_{ik}   / (\sum_{i'} p_{i'k})^{\frac{1}{2}}   $
	}
	\caption{\textit{DEPICT} Algorithm}
	\label{alg:1}
\end{algorithm}


\section{Experiments} \label{section:experiment}
In this section, we first evaluate \textit{DEPICT}\footnote{Our code is available in \url{https://github.com/herandy/DEPICT}} in comparison with state-of-the-art clustering methods on several benchmark image datasets. Then, the running speed of the best clustering models are compared. Moreover, we examine different learning approaches for training \textit{DEPICT}. Finally, we analyze the performance of \textit{DEPICT} model on semi-supervised classification tasks.

\noindent \textbf{Datasets}:
In order to show that \textit{DEPICT} works well with various kinds of datasets, we have chosen the following handwritten digit and face image datasets. Considering that clustering tasks are fully unsupervised, we concatenate the training and testing samples when applicable.
\textit{MNIST-full}: A dataset containing a total of 70,000 handwritten digits with 60,000 training and 10,000 testing samples, each being a 32 by 32 monochrome image \cite{lecun1998gradient}.
\textit{MNIST-test}: A dataset which only consists of the testing part of \textit{MNIST-full} data.
\textit{USPS}: It is a handwritten digits dataset from the \textit{USPS} postal service, containing 11,000 samples of 16 by 16 images.
\textit{CMU-PIE}: A dataset including 32 by 32 face images of 68 people with 4 different expressions \cite{sim2002cmu}.
\textit{Youtube-Face (YTF)}: Following \cite{yang2016joint}, we choose the first 41 subjects of YTF dataset. Faces inside images are first cropped and then resized to 55 by 55 sizes \cite{wolf2011face}.
\textit{FRGC}: Using the 20 random selected subjects in \cite{yang2016joint} from the original dataset, we collect 2,462 face images. Similarly, we first crop the face regions and resize them into 32 by 32 images.
Table \ref{table:datasets} provides a brief description of each dataset.

\begin{table}[h]
\centering
	\small
	\setlength{\tabcolsep}{7pt}
	\renewcommand{\arraystretch}{0.9}
	\begin{tabular}{l|c|c|c}
		\hline
		{Dataset}  & \# Samples 	& \# Classes 	& \# Dimensions \\ \hline \hline
		\textit{MNIST-full} 	& 70,000 		& 10		& 1$\times$28$\times$28 	  \\
		\textit{MNIST-test}	& 10,000 		& 10		& 1$\times$28$\times$28 	  \\
		\textit{USPS} 	& 11,000 		& 10		& 1$\times$16$\times$16 	  \\
		\textit{FRGC}	& 2,462 		& 20		& 3$\times$32$\times$32 	  \\
		\textit{YTF}	& 10,000 		& 41		& 3$\times$55$\times$55 	  \\
		\textit{CMU-PIE}	& 2,856 		& 68		& 1$\times$32$\times$32 	  \\
		
		\hline
	\end{tabular}
	\caption{Dataset Descriptions}
	\label{table:datasets}
\end{table}

\noindent \textbf{Clustering Metrics}:
We have used 2 of the most popular evaluation criteria widely used for clustering algorithms, accuracy (ACC) and normalized mutual information (NMI).
The best mapping between cluster assignments and true labels is computed using the Hungarian algorithm \cite{kuhn1955hungarian} to measure accuracy.
NMI calculates the normalized measure of similarity between two labels of the same data \cite{xu2003document}. Results of NMI do not change by permutations of clusters (classes), and they are normalized to have $[0,1]$ range, with $0$ meaning no correlation and $1$ exhibiting perfect correlation.

\begin{table*}[t]
	\small
	\setlength{\tabcolsep}{4pt}
	\renewcommand{\arraystretch}{0.95}
	\centering
	\begin{tabular}{l|cc|cc|cc|cc|cc|cc|c}
		\hline
		Dataset &
		\multicolumn{2}{c|}{\textit{MNIST-full}} &
		\multicolumn{2}{c|}{\textit{MNIST-test}} &
		\multicolumn{2}{c|}{\textit{USPS}} &
		\multicolumn{2}{c|}{\textit{FRGC}} &
		\multicolumn{2}{c|}{\textit{YTF}} &
		\multicolumn{2}{c|}{\textit{CMU-PIE}} &
		\multirow{2}{*}{\begin{tabular}{c}\# tuned \\ HPs\end{tabular}}\\ \cline{0-12}
		& NMI & ACC & NMI & ACC & NMI & ACC & NMI & ACC& NMI & ACC& NMI & ACC & \\
		\hline
		\textit{$K$-means} 	& 0.500$^{*}$ 	& 0.534$^{*}$ 	& 0.501$^{*}$ 	& 0.547$^{*}$ 	& 0.450$^{*}$ 	& 0.460$^{*}$ 	& 0.287$^{*}$ 	& 0.243$^{*}$ 	& 0.776$^{*}$ 	& 0.601$^{*}$ 	& 0.432$^{*}$ 	& 0.223$^{*}$ 	&	0	\\
		\textit{N-Cuts} 	& 0.411 		& 0.327 		& 0.753 		& 0.304 		& 0.675 		& 0.314 		& 0.285 		& 0.235 		& 0.742 		& 0.536 		& 0.411 		& 0.155 		&		0		\\
		\textit{SC-ST} 		& 0.416 		& 0.311 		& 0.756 		& 0.454 		& 0.726 		& 0.308 		& 0.431 		& 0.358			& 0.620			& 0.290 		& 0.581 		& 0.293 		&		0		\\
		\textit{SC-LS} 		& 0.706 		& 0.714		 	& 0.756 		& 0.740			& 0.681			& 0.659 		& 0.550 		& 0.407 		& 0.759 		& 0.544 		& 0.788 		& 0.549 		&		0		\\
		\textit{AC-GDL} 	& 0.017 		& 0.113 		& 0.844 		& 0.933			& 0.824			& 0.867 		& 0.351 		& 0.266 		& 0.622 		& 0.430			& 0.934 		& 0.842 		&		1		\\
		\textit{AC-PIC} 	& 0.017 		& 0.115 		& 0.853 		& 0.920			& 0.840			& 0.855 		& 0.415 		& 0.320 		& 0.697 		& 0.472 		& 0.902 		& 0.797 		&		0	\\
		\textit{SEC} 		& 0.779$^{*}$ 	& 0.804$^{*}$ 	& 	0.790$^{*}$	& 	0.815$^{*}$	& 0.511$^{*}$	& 0.544$^{*}$ 	&  	-			&  	-			&  	 -			&  	-			&	-			&	-			&	1	\\
		\textit{LDMGI} 		& 0.802$^{*}$ 	& 0.842$^{*}$ 	& 	0.811$^{*}$	& 	0.847$^{*}$	& 0.563$^{*}$	& 0.580$^{*}$	&  	-			&  	-			& 	 -			&	-			&	-			&	-			&	1	\\  \hline \hline
		\textit{NMF-D} 		& 0.152$^{*}$ 	& 0.175$^{*}$ 	& 0.241$^{*}$ 	& 0.250$^{*}$ 	& 0.287$^{*}$	& 0.382$^{*}$ 	& 0.259$^{*}$ 	& 0.274$^{*}$	& 0.562$^{*}$ 	& 0.536$^{*}$ 	& 0.920$^{*}$ 	& 0.810$^{*}$ 	& 0 \\
		\textit{TSC-D} 		& 0.651 		& 0.692			&  	-			& 	-			& 	-			&  	-			&   -			&  	-			&  	 -			& 	-			& 	-			&  - 			& 2	\\
		\textit{DEC} 		& 0.816$^{*}$ 	& 0.844$^{*}$ 	&  	0.827$^{*}$	& 	0.859$^{*}$	& 0.586$^{*}$ 	& 0.619$^{*}$	&  	0.505$^{*}$	&  	0.378$^{*}$	&  	 0.446$^{*}$	&  0.371$^{*}$	&	0.924$^{*}$	&	0.801$^{*}$ &	1	\\
		\textit{JULE-SF} 	& 0.906 		& 0.959 		& 0.876 		& 0.940			& 0.858			& 0.922 		& 0.566 		& 0.461 		 &\textbf{0.848} &\textbf{0.684} &0.984 			& 0.980 		&  3 \\
		\textit{JULE-RC} 	& 0.913 		& 0.964 		&\textbf{0.915} & 0.961 		& 0.913			& 0.950 		& 0.574 		& 0.461 		 &\textbf{0.848} &\textbf{0.684} & \textbf{1.00} & \textbf{1.00} & 3 \\ \hline \hline
		\textit{DEPICT}		&\textbf{0.917} &\textbf{0.965} &  \textbf{0.915} 		&\textbf{0.963} &\textbf{0.927} &\textbf{0.964} &\textbf{0.610} & \textbf{0.470}& 0.802 		& 0.621 		& 0.974 		& 0.883  		&	0 \\
		\hline
	\end{tabular}
	\vspace{-0.15cm}
	\caption{Clustering performance of different algorithms on image datasets based on accuracy (ACC) and normalized mutual information (NMI). The numbers of tuned hyper-parameters (\# tuned HPs) using the supervisory signals are also shown for each algorithm. The results of alternative models are reported from original papers, except the ones marked by (${*}$) on top, which are obtained by us running the released code. We put dash marks (-) for the results that are not practical to obtain. }
	\label{table:2}
	\vspace{-0.2cm}
\end{table*}

\subsection{Evaluation of Clustering Algorithm}

\noindent \textbf{Alternative Models}:
We compare our clustering model, \textit{DEPICT}, with several baseline and state-of-the-art clustering algorithms, including \textit{$K$-means}, normalized cuts (\textit{N-Cuts}) \cite{shi2000normalized}, self-tuning spectral clustering (\textit{SC-ST}) \cite{zelnik2004self}, large-scale spectral clustering (\textit{SC-LS}) \cite{chen2011large}, graph degree linkage-based agglomerative clustering (\textit{AC-GDL}) \cite{zhang2012graph}, agglomerative clustering via path integral (\textit{AC-PIC}) \cite{zhang2013agglomerative}, spectral embedded clustering (\textit{SEC}) \cite{nie2011spectral}, local discriminant models and global integration (\textit{LDMGI}) \cite{yang2010image}, \textit{NMF} with deep model (\textit{NMF-D}) \cite{trigeorgis2014deep}, task-specific clustering with deep model (\textit{TSC-D}) \cite{wang2016learning}, deep embedded clustering (\textit{DEC}) \cite{xie2016unsupervised}, and joint unsupervised learning (\textit{JULE}) \cite{yang2016joint}.

\noindent \textbf{Implementation Details}:
We use a common architecture for \textit{DEPICT} and avoid tuning any hyper-parameters using the labeled data in order to provide a practical algorithm for real-world clustering tasks. For all datasets, we consider two convolutional layers followed by a fully connected layer in encoder and decoder pathways. While for all convolutional layers, the feature map size is 50 and the kernel size is about $5 \times 5$, the dimension of the embedding subspace is set equal to the number of clusters in each dataset. We also pick the proper stride, padding and crop to have an output size of about $10 \times 10$ in the second convolutional layer. Inspired by \cite{radford2015unsupervised}, we consider leaky rectified (leaky RELU) non-linearity \cite{maas2013rectifier} as the activation function of convolutional and fully connected layers, except in the last layer of encoder and first layer of decoder, which have Tanh non-linearity functions. Consequently, we normalize the image intensities to be in the range of $[-1, 1]$.
Moreover, we set the learning rate and dropout to $10^{-4}$ and $0.1$ respectively, adopt adam as our optimization method with the default hyper-parameters $\beta_1 = 0.9$, $\beta_2 = 0.999$, $\epsilon=1e-08$ \cite{kingma2014adam}. The weights of convolutional and fully connected layers are all initialized by Xavier approach \cite{glorot2010understanding}.
Since the clustering assignments in the first iterations are random and not reliable for clustering loss, we first train \textit{DEPICT} without clustering loss function for a while, then initialize the clustering assignment $q_{ik}$ by clustering the embedding subspace features via simple algorithms like $K$-means or \textit{AC-PIC}.

\noindent \textbf{Quantitative Comparison}:
We run \textit{DEPICT} and other clustering methods on each dataset. We followed the implementation details for \textit{DEPICT} and report the average results from 5 runs. For the rest, we present the best reported results either from their original papers or from \cite{yang2016joint}. For unreported results on specific datasets, we run the released code with hyper-parameters mentioned in the original papers, these results are marked by (${*}$) on top.  But, when the code is not publicly available, or running the released code is not practical, we put dash marks (-) instead of the corresponding results.
Moreover, we mention the number of hyper-parameters that are tuned using supervisory signals (labeled data) for each algorithm. Note that this number only shows the quantity of hyper-parameters, which are set differently for various datasets for better performance.

Table \ref{table:2} reports the clustering metrics, normalized mutual information (NMI) and accuracy (ACC), of the algorithms on the aforementioned datasets. As shown, \textit{DEPICT} outperforms other algorithms on four datasets and achieves competitive results on the remaining two. It should be noted that we think hyper-parameter tuning using supervisory signals is not feasible in real-world clustering tasks, and hence \textit{DEPICT} is a significantly better clustering algorithm compared to the alternative models in practice. For example, \textit{DEC}, \textit{SEC}, and \textit{LDMGI} report their best results by tuning one hyper-parameter over nine different options, and \textit{JULE-SF} and \textit{JULE-RC} achieve their good performance by tweaking several hyper-parameters over various datasets. However, we do not tune any hyper-parameters for \textit{DEPICT} using the labeled data and only report the result with the same (default) hyper-parameters for all datasets.

\subsection{Running Time Comparison}
	
	In order to evaluate the efficiency of our clustering algorithm in dealing with large-scale and high dimensional data, we compare the running speed of \textit{DEPICT} with its competing algorithms, \textit{JULE-SF} and \textit{JULE-RC}.
	Moreover, the fast versions of \textit{\textit{JULE-SF}} and \textit{JULE-RC} are also evaluated. Note that JULE-SF(fast) and \textit{JULE-RC}(fast) both require tuning one extra hyper-parameter for each dataset to achieve results similar to the original \textit{JULE} algorithms in Table 2 \cite{yang2016joint}.
	 We run \textit{DEPICT} and the released code for \textit{JULE} algorithms\footnote{\url{https://github.com/jwyang/JULE-Torch}}  on a machine with one Titan X pascal GPU and a Xeon E5-2699 CPU. 

\begin{figure}[t]
\centering
		\begin{center}
			\includegraphics[trim=0mm 6mm 19.5mm 12mm, clip,  width=0.90\linewidth]{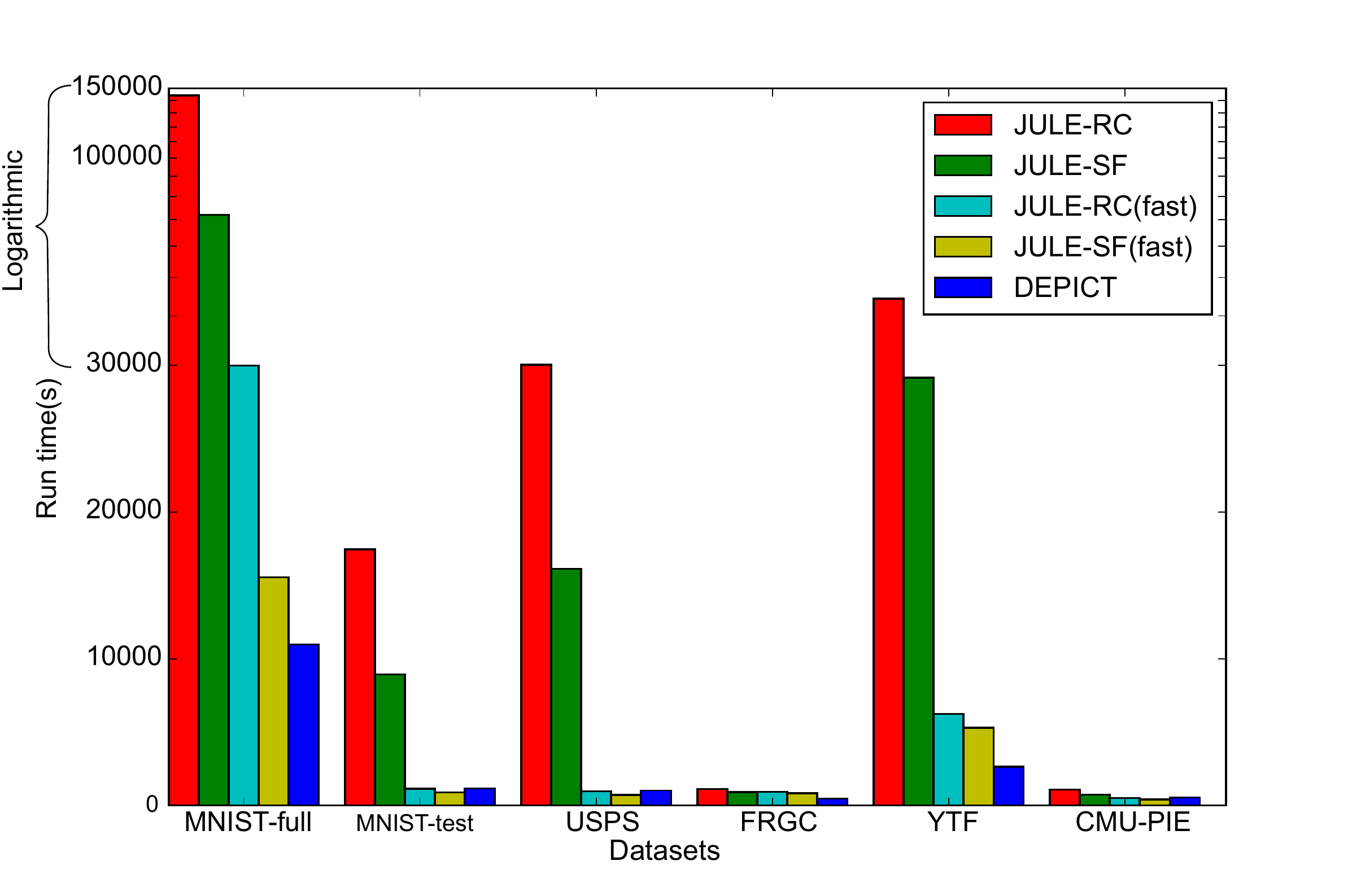}
		\end{center}
		\vspace{-0.5cm}
		\caption{Running time comparison of \textit{DEPICT} and \textit{JULE} clustering algorithms on image datasets.}
		\vspace{-0.4cm}
		\label{fig:5}
	\end{figure}

	Figure \ref{fig:5} illustrates the running time for \textit{DEPICT} and \textit{JULE} algorithms on all datasets. Note that running times of \textit{JULE-SF} and \textit{JULE-RC} are shown linearly from 0 to 30,000 and logarithmically for larger values for the sake of readability. In total, \textit{JULE-RC}, \textit{JULE-SF}, \textit{JULE-RC}(fast), \textit{JULE-SF}(fast) and \textit{DEPICT} take 66.1, 35.5, 11.0, 6.6 and 4.7 hours respectively to run over all datasets.  While all algorithms have approximately similar running times on small datasets (\textit{FRGC} and \textit{CMU-PIE}), when dealing with the large-scale and high-dimensional datasets (\textit{MNIST-full} and \textit{YTF}), \textit{DEPICT} almost shows a linear increase in the running time, but the running times of original JULE algorithms dramatically grow with the size and number of input data.  This outcome again emphasizes the practicality of \textit{DEPICT} for real-world clustering tasks.

\begin{table*}[t]
	\small
	\setlength{\tabcolsep}{5pt}
	\renewcommand{\arraystretch}{0.95}
	\centering
	\begin{tabular}{ll|ll|ll|ll|ll|ll|ll}
		\hline
		Dataset &
		&\multicolumn{2}{c|}{\textit{MNIST-full}} &
		\multicolumn{2}{c|}{\textit{MNIST-test}} &
		\multicolumn{2}{c|}{\textit{USPS}} &
		\multicolumn{2}{c|}{\textit{FRGC}} &
		\multicolumn{2}{c|}{\textit{YTF}} &
		\multicolumn{2}{c}{\textit{CMU-PIE}}\\ \hline
		&& NMI & ACC & NMI & ACC & NMI & ACC & NMI & ACC& NMI & ACC & NMI & ACC\\
		\hline
		\multirow{3}{*}{\begin{minipage}{2.2cm} \centering  \textit{Deep-ConvAE} + \hspace{1.5cm} \textit{AC-PIC} \end{minipage} }
		&	SdA	&	0.255	&	0.348	&	0.313	&	0.345	&	0.223	&	0.290	&	0.120	&	0.230	&	0.414	&	0.302	&	0.354	&	0.266	 \\
		&	RdA	&	0.615	&	0.455	&	0.859	&	0.900	&	0.886	&	0.866	&	0.443	&	0.363	&	0.597	&	0.425	&	0.912	&	0.817	 \\
		&	MdA	&	0.729	&	0.506	&	0.876	&	0.942	&	0.906	&	0.878	&	0.583	&	0.427	&	0.640	&	0.448	&	0.931	&	0.883	 \\
		\hline
		\hline
		\multirow{3}{*}{ \centering \textit{DEPICT}}
		&	SdA	&	0.365	&	0.427	&	0.353	&	0.390	&	0.328	&	0.412	&	0.211	&	0.300	&	0.414	&	0.302	&	0.354	&	0.266	 \\
		&	RdA	&	0.808	&	0.677	&	0.899	&	0.950	&	0.901	&	0.923	&	0.551	&	0.444	&	0.652	&	0.450	&	0.951	&	0.926	 \\
		& MdA  & \textbf{0.917} & \textbf{0.965} &  \textbf{0.915} & \textbf{0.963} & \textbf{0.927} & \textbf{0.964}  & \textbf{0.610} & \textbf{0.470} & \textbf{0.802} & \textbf{0.621} & \textbf{0.974} & \textbf{0.883}  \\
		\hline
	\end{tabular}
	\vspace{-0.25cm}
	\caption{Clustering performance of different learning approaches, including SdA, RdA and MdA, for training \textit{DEPICT} and \textit{Deep-ConvAE}+\textit{AC-PIC} models.}
	\label{table:3}
	\vspace{-0.35cm}
\end{table*}

\subsection{Evaluation of Learning Approach}
In order to evaluate our joint learning approach, we compare  several strategies for training \textit{DEPICT}. For training a multi-layer convolutional autoencoder, we analyze the following three approaches : 1) Standard stacked denoising autoencoder (SdA), in which the model is first pretrained using the reconstruction loss function in a layer-wise manner, and the encoder pathway is then fine-tuned using the clustering objective function \cite{vincent2010stacked}. 2) Another approach (RdA)  is suggested in \cite{xie2016unsupervised} to improve the SdA learning approach, in which all of the autoencoder layers are retrained after the pretraining step, only using the reconstruction of input layer while data is not corrupted by noise. The fine-tuning step is also done after the retraining step. 3) Our learning approach (MdA), in which the whole model is trained simultaneously using the joint reconstruction loss functions from all layers along with the clustering objective function.

Furthermore, we also examine the effect of clustering loss (through error back-prop) in constructing the embedding subspace. To do so, we train a similar multi-layer convolutional autoencoder (\textit{Deep-ConvAE}) only using the reconstruction loss function to generate the embedding subspace. Then, we run the best shallow clustering algorithm (\textit{AC-PIC}) on the embedded data. Hence, this model (\textit{Deep-ConvAE}+\textit{AC-PIC}) differs from \textit{DEPICT} in the sense that its embedding subspace is only constructed using the reconstruction loss and does not involve the clustering loss.

Table \ref{table:3} indicates the results of \textit{DEPICT} and \textit{Deep-ConvAE}+\textit{\textit{AC-PIC}} when using the different learning approaches. As expected, \textit{DEPICT} trained by our joint learning approach (MdA) consistently outperforms the other alternatives on all datasets. Interestingly, MdA learning approach shows promising results for \textit{Deep-ConvAE}+\textit{AC-PIC} model, where only reconstruction losses are used to train the embedding subspace. Thus, our learning approach is an efficient strategy for training autoencoder models due to its superior results and fast end-to-end training.

\subsection{Semi-Supervised Classification Performance}

	Representation learning in an unsupervised manner or using a small number of labeled data has recently attracted great attention. Due to the potential of our model in learning a discriminative embedding subspace, we evaluate \textit{DEPICT} in a semi-supervised classification task. Following the semi-supervised experiment settings \cite{rasmus2015semi,zhao2015stacked}, we train our model using a small random subset of  \textit{MNIST-training} dataset as labeled data and the remaining as unlabeled data. The classification error of \textit{DEPICT} is then computed using the \textit{MNIST-test} dataset, which is not seen during training.  Compared to our unsupervised learning approach, we only utilize the clusters corresponding to each labeled data in training process. In particular, only for labeled data, the cluster labels (assignments) are set using the best map technique from the original classification labels once, and then they will be fixed during the training step.
	
	Table \ref{table:5} shows the error results for several semi-supervised classification models using different numbers of labeled data. Surprisingly, \textit{DEPICT} achieves comparable results with the state-of-the-art, despite the fact that the semi-supervised classification models use 10,000 validation data to tune their hyper-parameters, \textit{DEPICT} only employs the labeled training data (e.g. 100) and does not tune any hyper-parameters. Although \textit{DEPICT} is not mainly designed for classification tasks, it outperforms several models including \textit{SWWAE} \cite{zhao2015stacked}, \textit{M1+M2} \cite{kingma2014semi}, and \textit{AtlasRBF} \cite{pitelis2014semi}, and has comparable results with the complicated \textit{Ladder} network \cite{rasmus2015semi}. These results further confirm the discriminative quality of the embedding features of \textit{DEPICT}.

\begin{table}[t]
	\small
	\setlength{\tabcolsep}{6pt}
	\renewcommand{\arraystretch}{0.90}
	\centering
	\begin{tabular}{l|ccc}
		\hline
		Model							& 100			& 1000 			& 3000 	\\\hline
		\textit{T-SVM} \cite{vapnik1998statistical}	& 16.81 		& 5.38			& 3.45	\\
		\textit{CAE} \cite{rifai2011contractive}		& 13.47			& 4.77			& 3.22	\\
		\textit{MTC} \cite{rifai2011manifold}		& 12.03 		& 3.64  		& 2.57	\\
		\textit{PL-DAE} \cite{lee2013pseudo}			& 10.49			& 3.46			& 2.69\\
		\textit{AtlasRBF} \cite{pitelis2014semi}		& 8.10			& 3.68			& - \\
		\textit{M1+M2} \cite{kingma2014semi}			& 3.33$\pm$0.14	& 2.40$\pm$0.05	& 2.18$\pm$0.04	\\
		\textit{SWWAE} \cite{zhao2015stacked}		& 8.71$\pm$0.34	& 2.83$\pm$0.10	& 2.10$\pm$0.22	\\
		\textit{Ladder} \cite{rasmus2015semi}		& \textbf{1.06$\pm$0.37}	& \textbf{0.84$\pm$0.08}	& 	-	\\ \hline
		\textit{DEPICT} 								& 2.65$\pm$0.35	& 2.10$\pm$0.11	& \textbf{1.91$\pm$0.06}	\\ \hline
	\end{tabular}
	\vspace{-0.15cm}
	\caption{Comparison of \textit{DEPICT} and several semi-supervised classification models in \textit{MNIST} dataset with different numbers of labeled data. }
	\label{table:5}
	\vspace{-0.4cm}
\end{table}

\section{Conclusion}\label{section:conclusion}
In this paper, we proposed a new deep clustering model, \textit{DEPICT}, consisting of a soft-max layer stacked on top of a multi-layer convolutional autoencoder. We employed a regularized relative entropy loss function for clustering, which leads to balanced cluster assignments. Adopting our autoencoder reconstruction loss function enhanced the embedding learning. Furthermore, a joint learning framework was introduced to train all network layers simultaneously and avoid layer-wise pretraining. Experimental results showed that \textit{DEPICT} is a good candidate for real-world clustering tasks, since it achieved superior or competitive results compared to alternative methods while having faster running speed and not needing hyper-parameter tuning. Efficiency of our joint learning approach was also confirmed in clustering and semi-supervised classification tasks.

{\small
	\bibliographystyle{ieee}
	\bibliography{myBib}

\begin{thebibliography}{10}\itemsep=-1pt

\bibitem{agrawal1998automatic}
R.~Agrawal, J.~Gehrke, D.~Gunopulos, and P.~Raghavan.
\newblock {\em Automatic subspace clustering of high dimensional data for data
  mining applications}, volume~27.
\newblock ACM, 1998.

\bibitem{bahmani2012scalable}
B.~Bahmani, B.~Moseley, A.~Vattani, R.~Kumar, and S.~Vassilvitskii.
\newblock Scalable k-means++.
\newblock {\em Proceedings of the VLDB Endowment}, 5(7):622--633, 2012.

\bibitem{barber2005kernelized}
D.~Barber and F.~V. Agakov.
\newblock Kernelized infomax clustering.
\newblock In {\em Advances in neural information processing systems (NIPS)},
  pages 17--24, 2005.

\bibitem{bezdek1984fcm}
J.~C. Bezdek, R.~Ehrlich, and W.~Full.
\newblock Fcm: The fuzzy c-means clustering algorithm.
\newblock {\em Computers \& Geosciences}, 10(2-3):191--203, 1984.

\bibitem{biernacki2000assessing}
C.~Biernacki, G.~Celeux, and G.~Govaert.
\newblock Assessing a mixture model for clustering with the integrated
  completed likelihood.
\newblock {\em IEEE transactions on pattern analysis and machine intelligence},
  22(7):719--725, 2000.

\bibitem{chen2011large}
X.~Chen and D.~Cai.
\newblock Large scale spectral clustering with landmark-based representation.
\newblock In {\em AAAI}, 2011.

\bibitem{Huang2015ICCV}
H.~Gao, F.~Nie, X.~Li, and H.~Huang.
\newblock Multi-view subspace clustering.
\newblock {\em International Conference on Computer Vision (ICCV 2015)}, pages
  4238--4246, 2015.

\bibitem{Huang2016ICDM1}
H.~Gao, X.~Wang, and H.~Huang.
\newblock New robust clustering model for identifying cancer genome landscapes.
\newblock {\em IEEE International Conference on Data Mining (ICDM 2016)}, pages
  151--160, 2016.

\bibitem{glorot2010understanding}
X.~Glorot and Y.~Bengio.
\newblock Understanding the difficulty of training deep feedforward neural
  networks.
\newblock In {\em Aistats}, volume~9, pages 249--256, 2010.

\bibitem{heller2005bayesian}
K.~A. Heller and Z.~Ghahramani.
\newblock Bayesian hierarchical clustering.
\newblock In {\em Proceedings of the 22nd international conference on Machine
  learning (ICML)}. ACM, 2005.

\bibitem{kailing2004density}
K.~Kailing, H.-P. Kriegel, and P.~Kr{\"o}ger.
\newblock Density-connected subspace clustering for high-dimensional data.
\newblock In {\em Proceedings of the 2004 SIAM International Conference on Data
  Mining}, pages 246--256. SIAM, 2004.

\bibitem{keogh2001online}
E.~Keogh, S.~Chu, D.~Hart, and M.~Pazzani.
\newblock An online algorithm for segmenting time series.
\newblock In {\em Data Mining, 2001. ICDM 2001, Proceedings IEEE International
  Conference on}, pages 289--296. IEEE, 2001.

\bibitem{kingma2014adam}
D.~Kingma and J.~Ba.
\newblock Adam: A method for stochastic optimization.
\newblock {\em arXiv preprint arXiv:1412.6980}, 2014.

\bibitem{kingma2014semi}
D.~P. Kingma, S.~Mohamed, D.~J. Rezende, and M.~Welling.
\newblock Semi-supervised learning with deep generative models.
\newblock In {\em Advances in Neural Information Processing Systems (NIPS)},
  pages 3581--3589, 2014.

\bibitem{krause2010discriminative}
A.~Krause, P.~Perona, and R.~G. Gomes.
\newblock Discriminative clustering by regularized information maximization.
\newblock In {\em Advances in neural information processing systems (NIPS)},
  pages 775--783, 2010.

\bibitem{kuhn1955hungarian}
H.~W. Kuhn.
\newblock The hungarian method for the assignment problem.
\newblock {\em Naval research logistics quarterly}, 2(1-2):83--97, 1955.

\bibitem{lecun1998gradient}
Y.~LeCun, L.~Bottou, Y.~Bengio, and P.~Haffner.
\newblock Gradient-based learning applied to document recognition.
\newblock {\em Proceedings of the IEEE}, 86(11):2278--2324, 1998.

\bibitem{lee2013pseudo}
D.-H. Lee.
\newblock Pseudo-label: The simple and efficient semi-supervised learning
  method for deep neural networks.
\newblock In {\em Workshop on Challenges in Representation Learning, ICML},
  volume~3, page~2, 2013.

\bibitem{li2004minimum}
H.~Li, K.~Zhang, and T.~Jiang.
\newblock Minimum entropy clustering and applications to gene expression
  analysis.
\newblock In {\em Computational Systems Bioinformatics Conference, 2004. CSB
  2004. Proceedings. 2004 IEEE}, pages 142--151. IEEE, 2004.

\bibitem{Huang2015AAAI2}
Y.~Li, F.~Nie, H.~Huang, and J.~Huang.
\newblock Large-scale multi-view spectral clustering via bipartite graph.
\newblock {\em Twenty-Ninth AAAI Conference on Artificial Intelligence (AAAI
  2015)}, 2015.

\bibitem{lloyd1982least}
S.~Lloyd.
\newblock Least squares quantization in pcm.
\newblock {\em IEEE transactions on information theory}, 28(2):129--137, 1982.

\bibitem{Dijun:ICDE10}
D.~Luo, C.~Ding, and H.~Huang.
\newblock Consensus spectral clustering.
\newblock {\em ICDE}, 2010.

\bibitem{maas2013rectifier}
A.~L. Maas, A.~Y. Hannun, and A.~Y. Ng.
\newblock Rectifier nonlinearities improve neural network acoustic models.
\newblock In {\em Proc. ICML}, volume~30, 2013.

\bibitem{nesterov2013introductory}
Y.~Nesterov.
\newblock {\em Introductory lectures on convex optimization: A basic course},
  volume~87.
\newblock Springer Science \& Business Media, 2013.

\bibitem{ng2002spectral}
A.~Y. Ng, M.~I. Jordan, Y.~Weiss, et~al.
\newblock On spectral clustering: Analysis and an algorithm.
\newblock {\em Advances in neural information processing systems (NIPS)},
  2:849--856, 2002.

\bibitem{Huang2016AAAI5}
F.~Nie, C.~Deng, H.~Wang, X.~Gao, and H.~Huang.
\newblock New l1-norm relaxations and optimizations for graph clustering.
\newblock {\em Thirtieth AAAI Conference on Artificial Intelligence (AAAI
  2016)}, 2016.

\bibitem{Huang2016IJCAI1}
F.~Nie and H.~Huang.
\newblock Subspace clustering via new discrete group structure constrained
  low-rank model.
\newblock {\em 25th International Joint Conference on Artificial Intelligence
  (IJCAI)}, pages 1874--1880, 2016.

\bibitem{Huang2014KDD2}
F.~Nie, X.~Wang, and H.~Huang.
\newblock Clustering and projected clustering via adaptive neighbor assignment.
\newblock {\em The 20th ACM SIGKDD Conference on Knowledge Discovery and Data
  Mining (KDD 2014)}, pages 977--986, 2014.

\bibitem{nie2016constrained}
F.~Nie, X.~Wang, M.~I. Jordan, and H.~Huang.
\newblock The constrained laplacian rank algorithm for graph-based clustering.
\newblock In {\em AAAI}, pages 1969--1976. Citeseer, 2016.

\bibitem{nie2011spectral}
F.~Nie, Z.~Zeng, I.~W. Tsang, D.~Xu, and C.~Zhang.
\newblock Spectral embedded clustering: A framework for in-sample and
  out-of-sample spectral clustering.
\newblock {\em IEEE Transactions on Neural Networks}, 22(11):1796--1808, 2011.

\bibitem{pitelis2014semi}
N.~Pitelis, C.~Russell, and L.~Agapito.
\newblock Semi-supervised learning using an unsupervised atlas.
\newblock In {\em Joint European Conference on Machine Learning and Knowledge
  Discovery in Databases}, pages 565--580. Springer, 2014.

\bibitem{radford2015unsupervised}
A.~Radford, L.~Metz, and S.~Chintala.
\newblock Unsupervised representation learning with deep convolutional
  generative adversarial networks.
\newblock {\em arXiv preprint arXiv:1511.06434}, 2015.

\bibitem{raina2003classification}
R.~Raina, Y.~Shen, A.~Y. Ng, and A.~McCallum.
\newblock Classification with hybrid generative/discriminative models.
\newblock In {\em In Advances in Neural Information Processing Systems (NIPS)},
  volume~16, 2003.

\bibitem{rasmus2015semi}
A.~Rasmus, M.~Berglund, M.~Honkala, H.~Valpola, and T.~Raiko.
\newblock Semi-supervised learning with ladder networks.
\newblock In {\em Advances in Neural Information Processing Systems (NIPS)},
  pages 3546--3554, 2015.

\bibitem{rifai2011manifold}
S.~Rifai, Y.~N. Dauphin, P.~Vincent, Y.~Bengio, and X.~Muller.
\newblock The manifold tangent classifier.
\newblock In {\em Advances in Neural Information Processing Systems (NIPS)},
  pages 2294--2302, 2011.

\bibitem{rifai2011contractive}
S.~Rifai, P.~Vincent, X.~Muller, X.~Glorot, and Y.~Bengio.
\newblock Contractive auto-encoders: Explicit invariance during feature
  extraction.
\newblock In {\em Proceedings of the 28th international conference on machine
  learning (ICML-11)}, pages 833--840, 2011.

\bibitem{roth2003feature}
V.~Roth and T.~Lange.
\newblock Feature selection in clustering problems.
\newblock In {\em In Advances in Neural Information Processing Systems (NIPS)},
  pages 473--480, 2003.

\bibitem{sadoughi2015retrieving}
N.~Sadoughi and C.~Busso.
\newblock Retrieving target gestures toward speech driven animation with
  meaningful behaviors.
\newblock In {\em Proceedings of the 2015 ACM on International Conference on
  Multimodal Interaction}, pages 115--122. ACM, 2015.

\bibitem{sadoughi2015msp}
N.~Sadoughi, Y.~Liu, and C.~Busso.
\newblock Msp-avatar corpus: Motion capture recordings to study the role of
  discourse functions in the design of intelligent virtual agents.
\newblock In {\em Automatic Face and Gesture Recognition (FG), 2015 11th IEEE
  International Conference and Workshops on}, 2015.

\bibitem{sargin2008analysis}
M.~E. Sargin, Y.~Yemez, E.~Erzin, and A.~M. Tekalp.
\newblock Analysis of head gesture and prosody patterns for prosody-driven
  head-gesture animation.
\newblock {\em IEEE Transactions on Pattern Analysis and Machine Intelligence},
  30(8):1330--1345, 2008.

\bibitem{shi2000normalized}
J.~Shi and J.~Malik.
\newblock Normalized cuts and image segmentation.
\newblock {\em IEEE Transactions on pattern analysis and machine intelligence},
  22(8):888--905, 2000.

\bibitem{shinnou2008spectral}
H.~Shinnou and M.~Sasaki.
\newblock Spectral clustering for a large data set by reducing the similarity
  matrix size.
\newblock In {\em LREC}, 2008.

\bibitem{sim2002cmu}
T.~Sim, S.~Baker, and M.~Bsat.
\newblock The cmu pose, illumination, and expression (pie) database.
\newblock In {\em Automatic Face and Gesture Recognition, 2002. Proceedings.
  Fifth IEEE International Conference on}, pages 46--51. IEEE, 2002.

\bibitem{srivastava2014dropout}
N.~Srivastava, G.~E. Hinton, A.~Krizhevsky, I.~Sutskever, and R.~Salakhutdinov.
\newblock Dropout: a simple way to prevent neural networks from overfitting.
\newblock {\em Journal of Machine Learning Research}, 15(1):1929--1958, 2014.

\bibitem{tian2014learning}
F.~Tian, B.~Gao, Q.~Cui, E.~Chen, and T.-Y. Liu.
\newblock Learning deep representations for graph clustering.
\newblock In {\em AAAI}, pages 1293--1299, 2014.

\bibitem{trigeorgis2014deep}
G.~Trigeorgis, K.~Bousmalis, S.~Zafeiriou, and B.~Schuller.
\newblock A deep semi-nmf model for learning hidden representations.
\newblock In {\em ICML}, pages 1692--1700, 2014.

\bibitem{vapnik1998statistical}
V.~Vapnik.
\newblock {\em Statistical learning theory}, volume~1.
\newblock Wiley New York, 1998.

\bibitem{vincent2010stacked}
P.~Vincent, H.~Larochelle, I.~Lajoie, Y.~Bengio, and P.-A. Manzagol.
\newblock Stacked denoising autoencoders: Learning useful representations in a
  deep network with a local denoising criterion.
\newblock {\em Journal of Machine Learning Research}, 11(Dec):3371--3408, 2010.

\bibitem{Huang2014ECML1}
D.~Wang, F.~Nie, and H.~Huang.
\newblock Unsupervised feature selection via unified trace ratio formulation
  and k-means clustering (track).
\newblock {\em European Conference on Machine Learning and Principles and
  Practice of Knowledge Discovery in Databases (ECML PKDD 2014)}, pages
  306--321, 2014.

\bibitem{Huang2013ICML1}
H.~Wang, F.~Nie, and H.~Huang.
\newblock Multi-view clustering and feature learning via structured sparsity.
\newblock {\em The 30th International Conference on Machine Learning (ICML
  2013)}, pages 352--360, 2013.

\bibitem{wang2016structured}
X.~Wang, F.~Nie, and H.~Huang.
\newblock Structured doubly stochastic matrix for graph based clustering:
  Structured doubly stochastic matrix.
\newblock In {\em ACM SIGKDD International Conference on Knowledge Discovery
  and Data Mining}, pages 1245--1254, 2016.

\bibitem{wang2016learning}
Z.~Wang, S.~Chang, J.~Zhou, M.~Wang, and T.~S. Huang.
\newblock Learning a task-specific deep architecture for clustering.
\newblock In {\em Proceedings of the 2016 SIAM International Conference on Data
  Mining}, pages 369--377. SIAM, 2016.

\bibitem{wang2015joint}
Z.~Wang, Y.~Yang, S.~Chang, J.~Li, S.~Fong, and T.~S. Huang.
\newblock A joint optimization framework of sparse coding and discriminative
  clustering.
\newblock In {\em International Joint Conference on Artificial Intelligence
  (IJCAI)}, volume~1, page~4, 2015.

\bibitem{williams1999mcmc}
C.~K. Williams.
\newblock A mcmc approach to hierarchical mixture modelling.
\newblock In {\em Advances in neural information processing systems (NIPS)},
  pages 680--686, 1999.

\bibitem{wolf2011face}
L.~Wolf, T.~Hassner, and I.~Maoz.
\newblock Face recognition in unconstrained videos with matched background
  similarity.
\newblock In {\em Computer Vision and Pattern Recognition (CVPR), 2011 IEEE
  Conference on}, pages 529--534. IEEE, 2011.

\bibitem{xie2016unsupervised}
J.~Xie, R.~Girshick, and A.~Farhadi.
\newblock Unsupervised deep embedding for clustering analysis.
\newblock In {\em International Conference on Machine Learning (ICML)}, 2016.

\bibitem{xie2015integrating}
P.~Xie and E.~P. Xing.
\newblock Integrating image clustering and codebook learning.
\newblock In {\em AAAI}, pages 1903--1909, 2015.

\bibitem{xu2004maximum}
L.~Xu, J.~Neufeld, B.~Larson, and D.~Schuurmans.
\newblock Maximum margin clustering.
\newblock In {\em Advances in neural information processing systems (NIPS)},
  pages 1537--1544, 2004.

\bibitem{xu2003document}
W.~Xu, X.~Liu, and Y.~Gong.
\newblock Document clustering based on non-negative matrix factorization.
\newblock In {\em Proceedings of the 26th annual international ACM SIGIR
  conference on Research and development in informaion retrieval}, pages
  267--273. ACM, 2003.

\bibitem{yang2016joint}
J.~Yang, D.~Parikh, and D.~Batra.
\newblock Joint unsupervised learning of deep representations and image
  clusters.
\newblock In {\em Proceedings of the IEEE Conference on Computer Vision and
  Pattern Recognition}, pages 5147--5156, 2016.

\bibitem{yang2010image}
Y.~Yang, D.~Xu, F.~Nie, S.~Yan, and Y.~Zhuang.
\newblock Image clustering using local discriminant models and global
  integration.
\newblock {\em IEEE Transactions on Image Processing}, 19(10):2761--2773, 2010.

\bibitem{ye2008discriminative}
J.~Ye, Z.~Zhao, and M.~Wu.
\newblock Discriminative k-means for clustering.
\newblock In {\em Advances in neural information processing systems (NIPS)},
  pages 1649--1656, 2008.

\bibitem{yeh2016semantic}
R.~Yeh, C.~Chen, T.~Y. Lim, M.~Hasegawa-Johnson, and M.~N. Do.
\newblock Semantic image inpainting with perceptual and contextual losses.
\newblock {\em arXiv preprint arXiv:1607.07539}, 2016.

\bibitem{zelnik2004self}
L.~Zelnik-Manor and P.~Perona.
\newblock Self-tuning spectral clustering.
\newblock In {\em In Advances in Neural Information Processing Systems (NIPS)},
  volume~17, page~16, 2004.

\bibitem{zhang2012graph}
W.~Zhang, X.~Wang, D.~Zhao, and X.~Tang.
\newblock Graph degree linkage: Agglomerative clustering on a directed graph.
\newblock In {\em European Conference on Computer Vision}, pages 428--441.
  Springer, 2012.

\bibitem{zhang2013agglomerative}
W.~Zhang, D.~Zhao, and X.~Wang.
\newblock Agglomerative clustering via maximum incremental path integral.
\newblock {\em Pattern Recognition}, 46(11):3056--3065, 2013.

\bibitem{zhao2008efficient}
B.~Zhao, F.~Wang, and C.~Zhang.
\newblock Efficient multiclass maximum margin clustering.
\newblock In {\em ICML}, pages 1248--1255, 2008.

\bibitem{zhao2015stacked}
J.~Zhao, M.~Mathieu, R.~Goroshin, and Y.~Lecun.
\newblock Stacked what-where auto-encoders.
\newblock {\em arXiv preprint arXiv:1506.02351}, 2015.

\bibitem{zhou2013hierarchical}
F.~Zhou, F.~De~la Torre, and J.~K. Hodgins.
\newblock Hierarchical aligned cluster analysis for temporal clustering of
  human motion.
\newblock {\em IEEE Transactions on Pattern Analysis and Machine Intelligence},
  35(3):582--596, 2013.

\end{thebibliography}
}

\begin{appendices}

\begin{figure*}[!b]
		\begin{subfigure}{.35\textwidth}
			\centering
			\includegraphics[trim=30mm 20mm 30mm 0mm, clip,  width=1\linewidth]{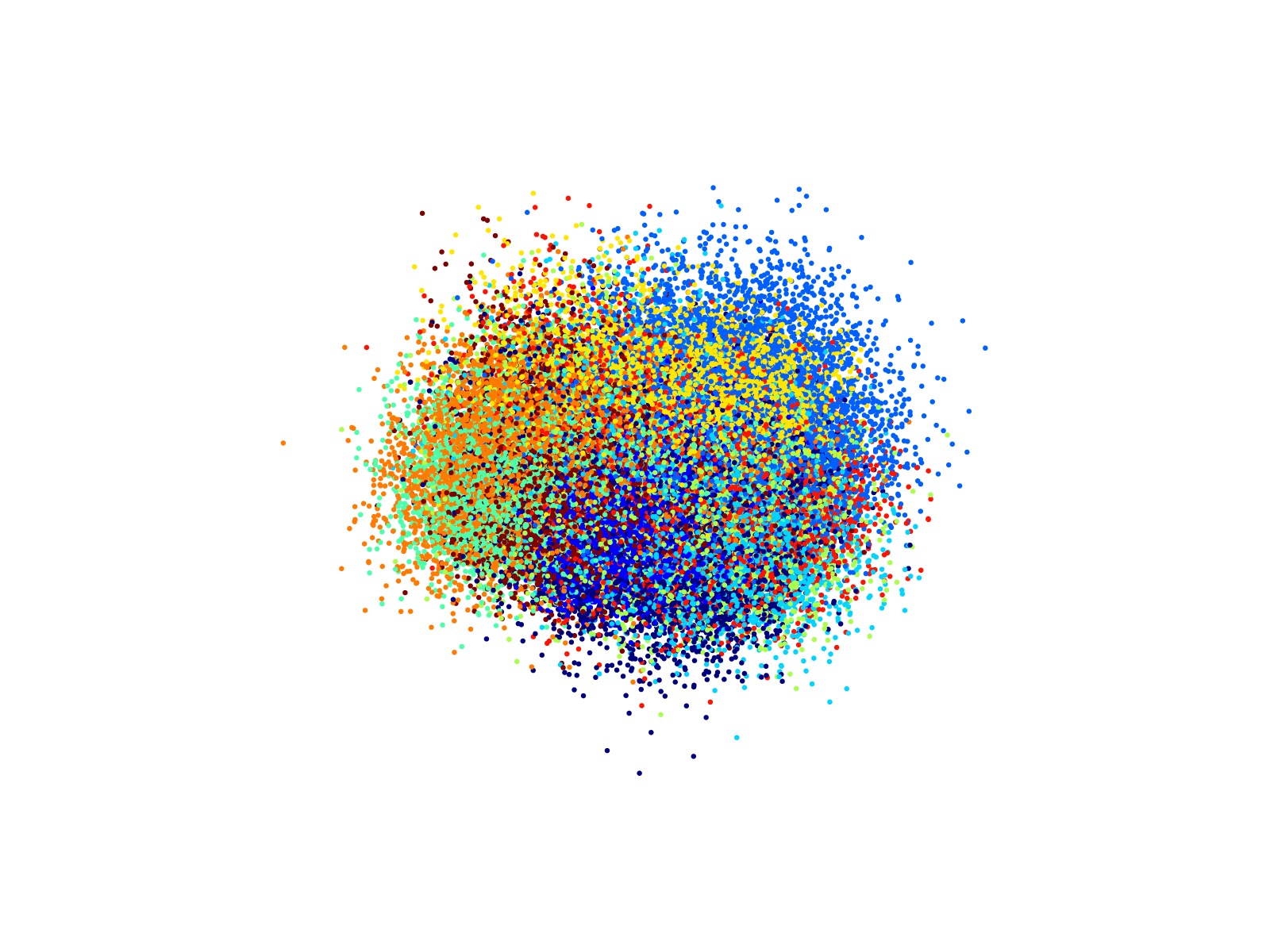}
			\caption{Initial stage on \textit{MNIST-full}}
		\end{subfigure}%
		\begin{subfigure}{.35\textwidth}
			\centering
			\includegraphics[trim=30mm 20mm 30mm 0mm, clip,  width=1\linewidth]{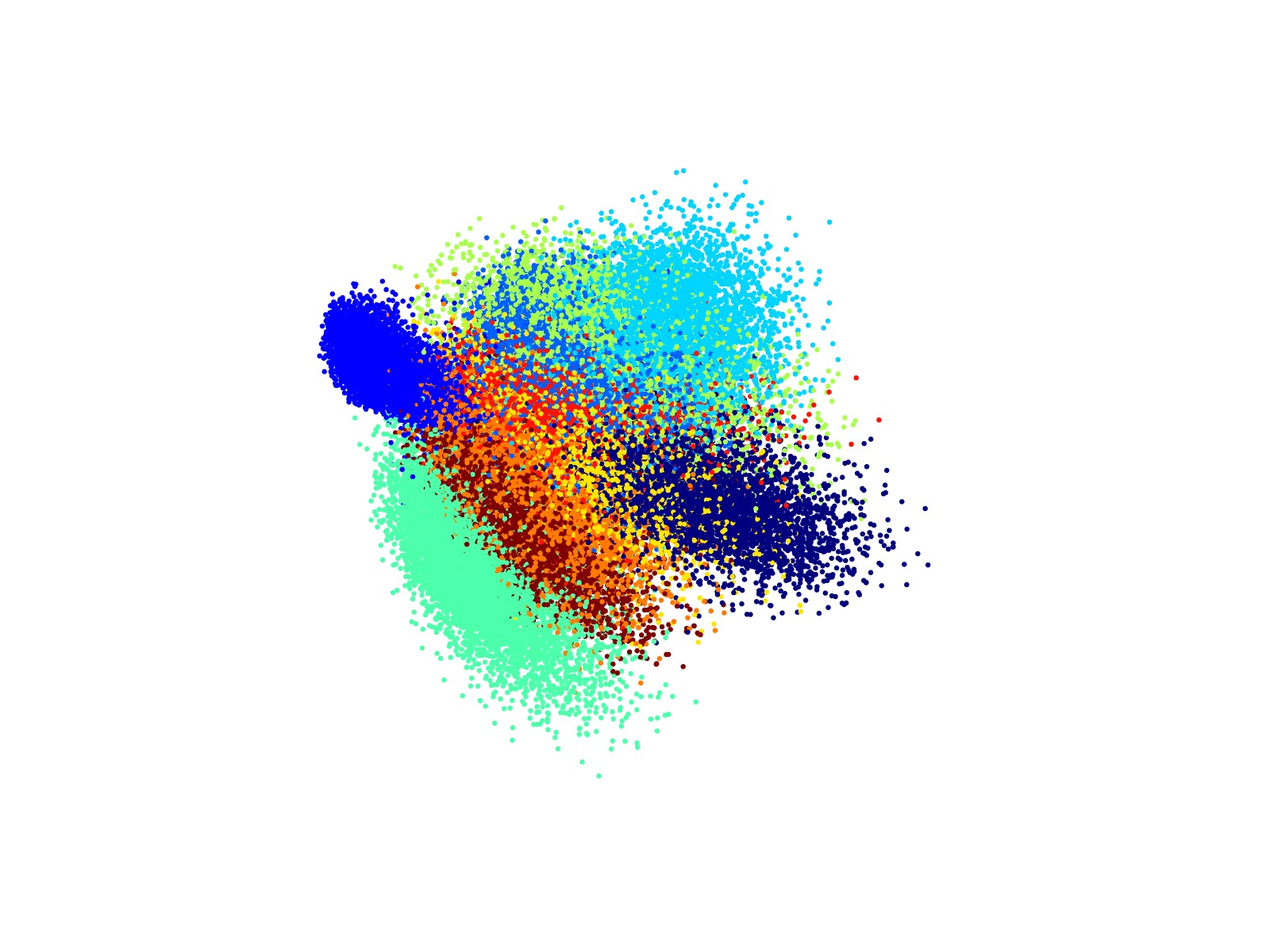}
			\caption{Intermediate stage on \textit{MNIST-full}}
		\end{subfigure}
		\begin{subfigure}{.35\textwidth}
			\centering
			\includegraphics[trim=30mm 20mm 30mm 0mm, clip,  width=1\linewidth]{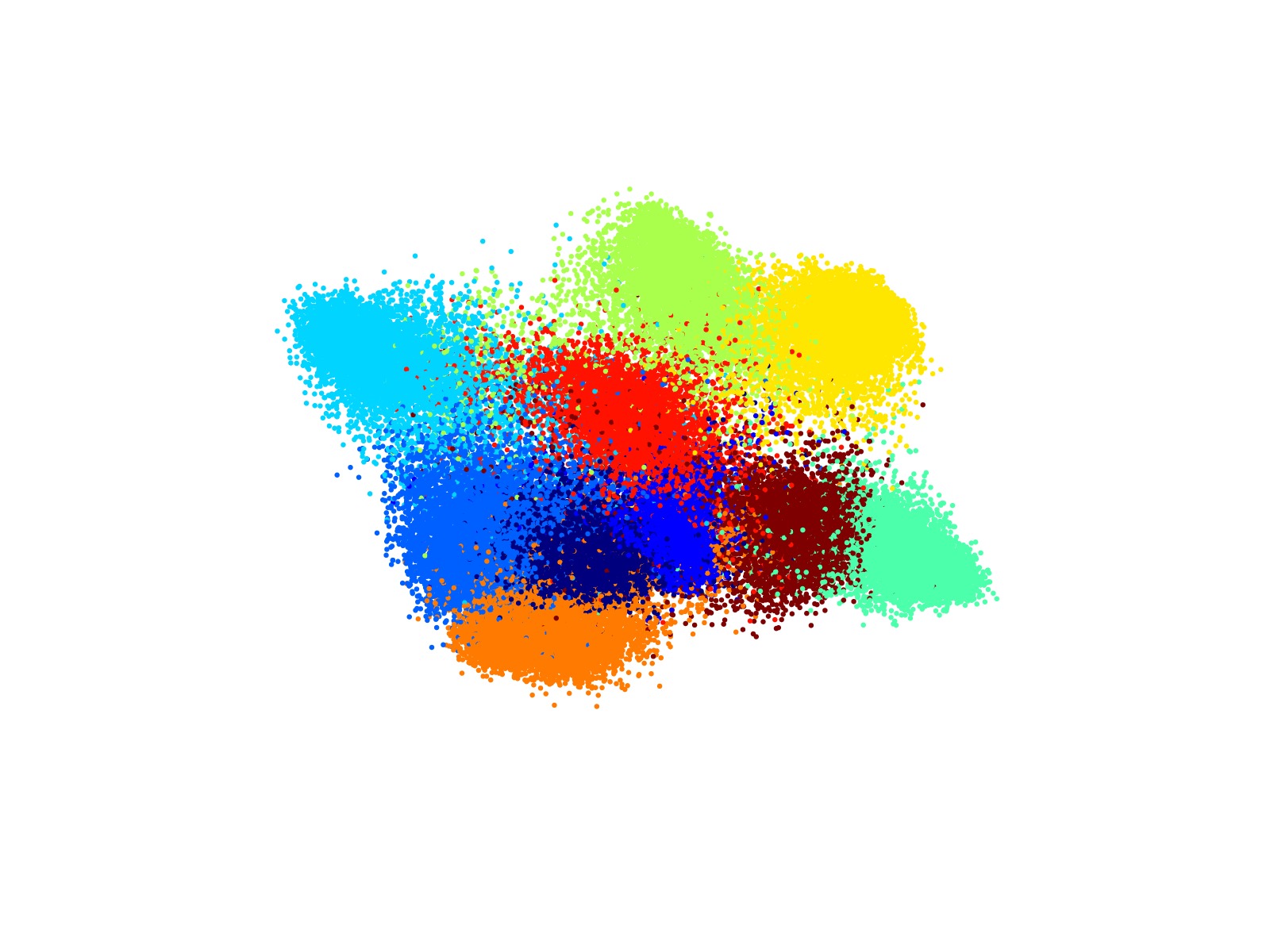}
			\caption{Final stage on \textit{MNIST-full}}
		\end{subfigure}
		\begin{subfigure}{.35\textwidth}
			\centering
			\includegraphics[trim=30mm 20mm 30mm 0mm, clip,  width=1\linewidth]{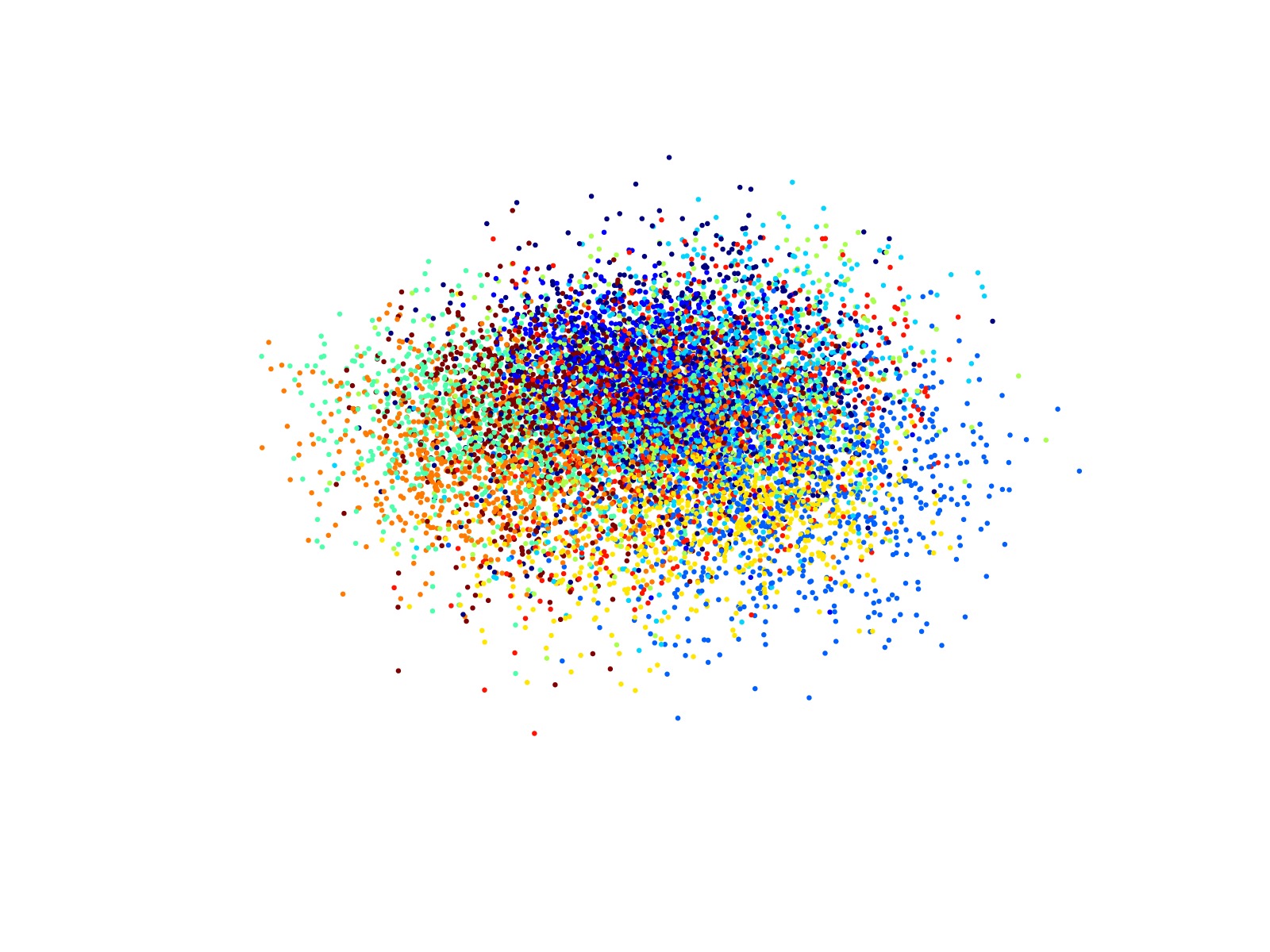}
			\caption{Initial stage on \textit{MNIST-test}}
		\end{subfigure}%
		\begin{subfigure}{.35\textwidth}
			\centering
			\includegraphics[trim=30mm 20mm 30mm 0mm, clip,  width=1\linewidth]{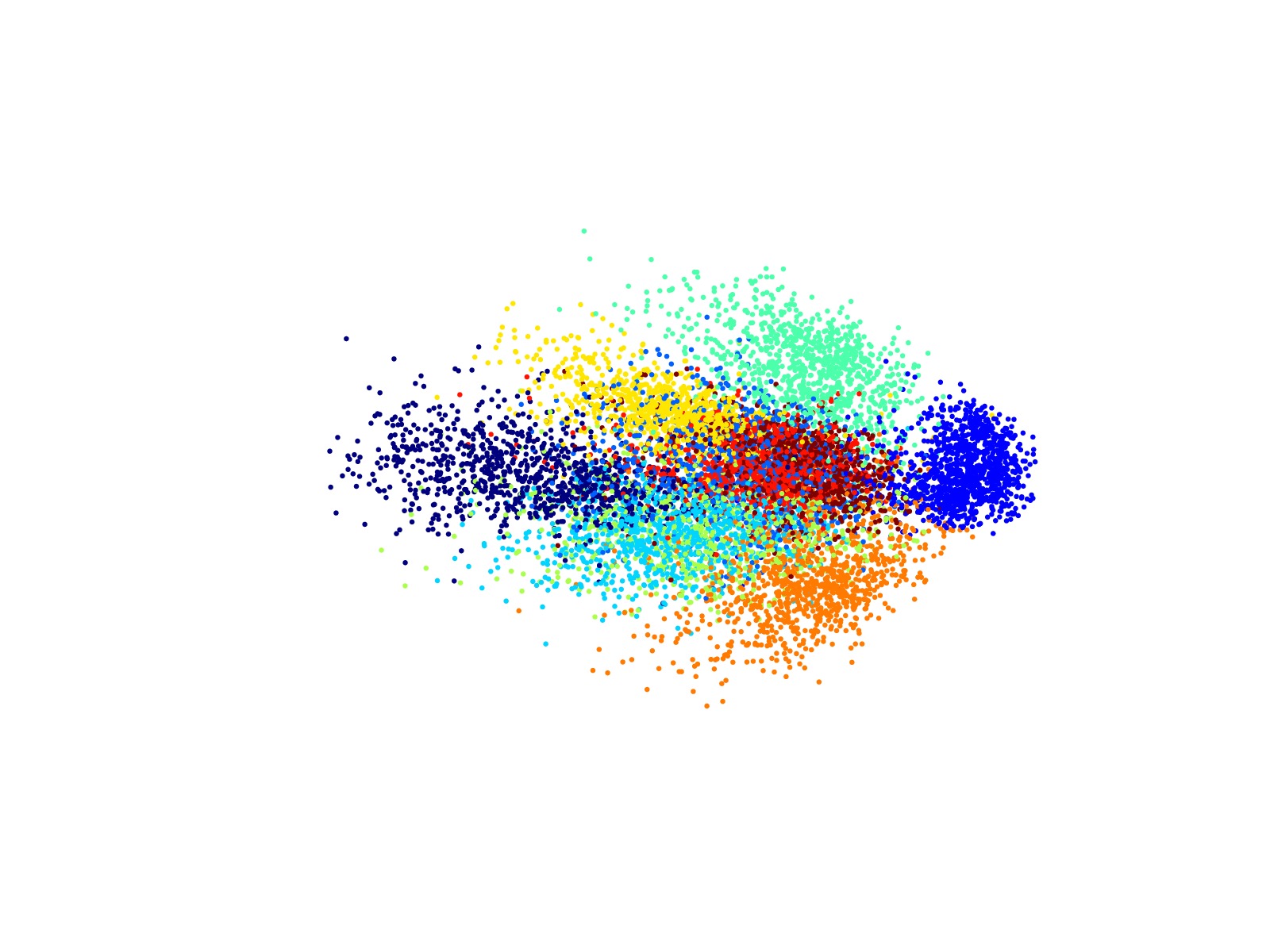}
			\caption{Intermediate stage on \textit{MNIST-test}}
		\end{subfigure}
		\begin{subfigure}{.35\textwidth}
			\centering
			\includegraphics[trim=30mm 20mm 30mm 0mm, clip,  width=1\linewidth]{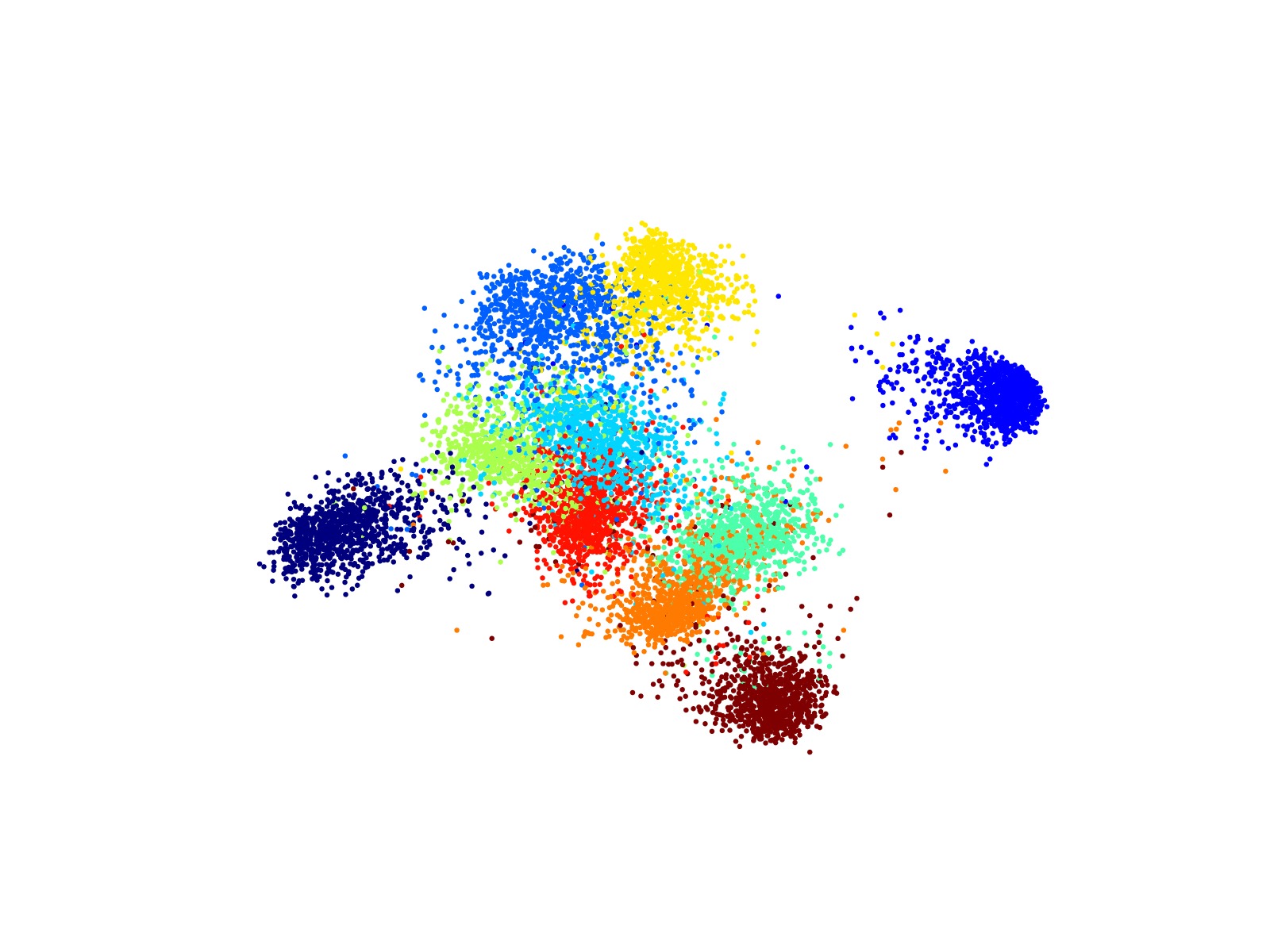}
			\caption{Final stage on \textit{MNIST-test}}
		\end{subfigure}
		\begin{subfigure}{.35\textwidth}
			\centering
			\includegraphics[trim=30mm 20mm 30mm 0mm, clip,  width=1\linewidth]{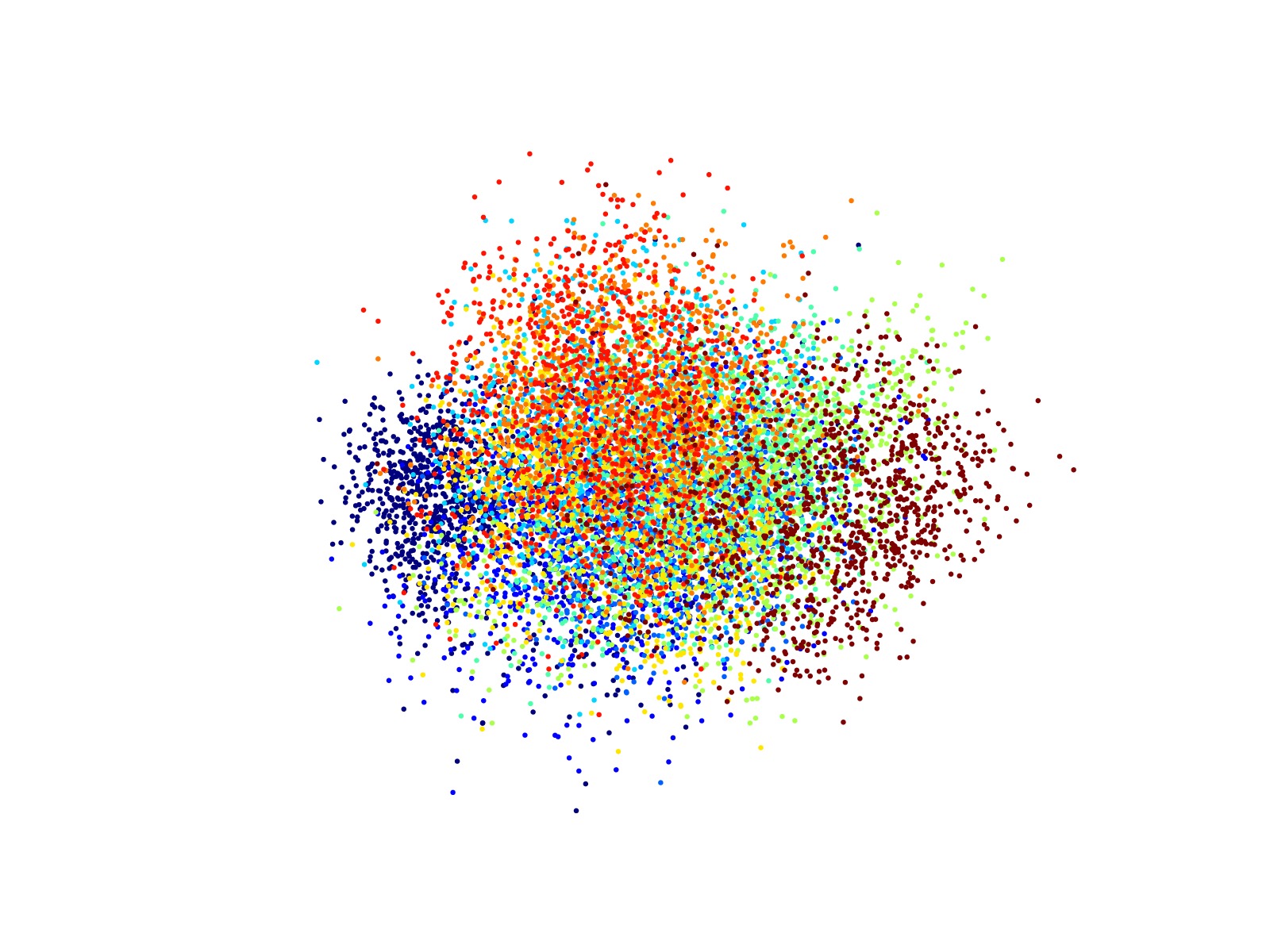}
			\caption{Initial stage on \textit{USPS}}
		\end{subfigure}%
		\begin{subfigure}{.35\textwidth}
			\centering
			\includegraphics[trim=30mm 20mm 30mm 0mm, clip,  width=1\linewidth]{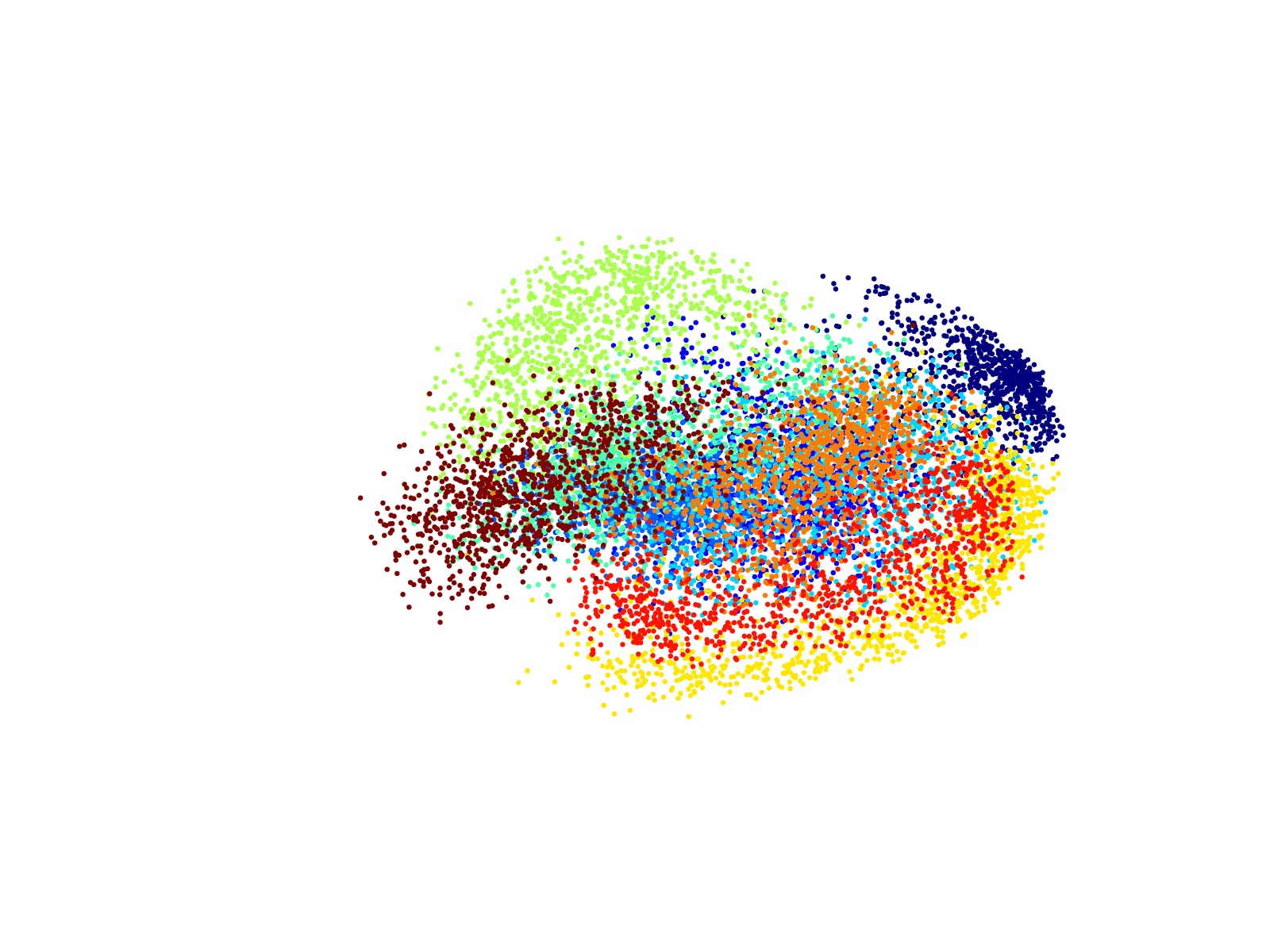}
			\caption{Intermediate stage on \textit{USPS}}
		\end{subfigure}
		\begin{subfigure}{.35\textwidth}
			\centering
			\includegraphics[trim=30mm 20mm 30mm 0mm, clip,  width=1\linewidth]{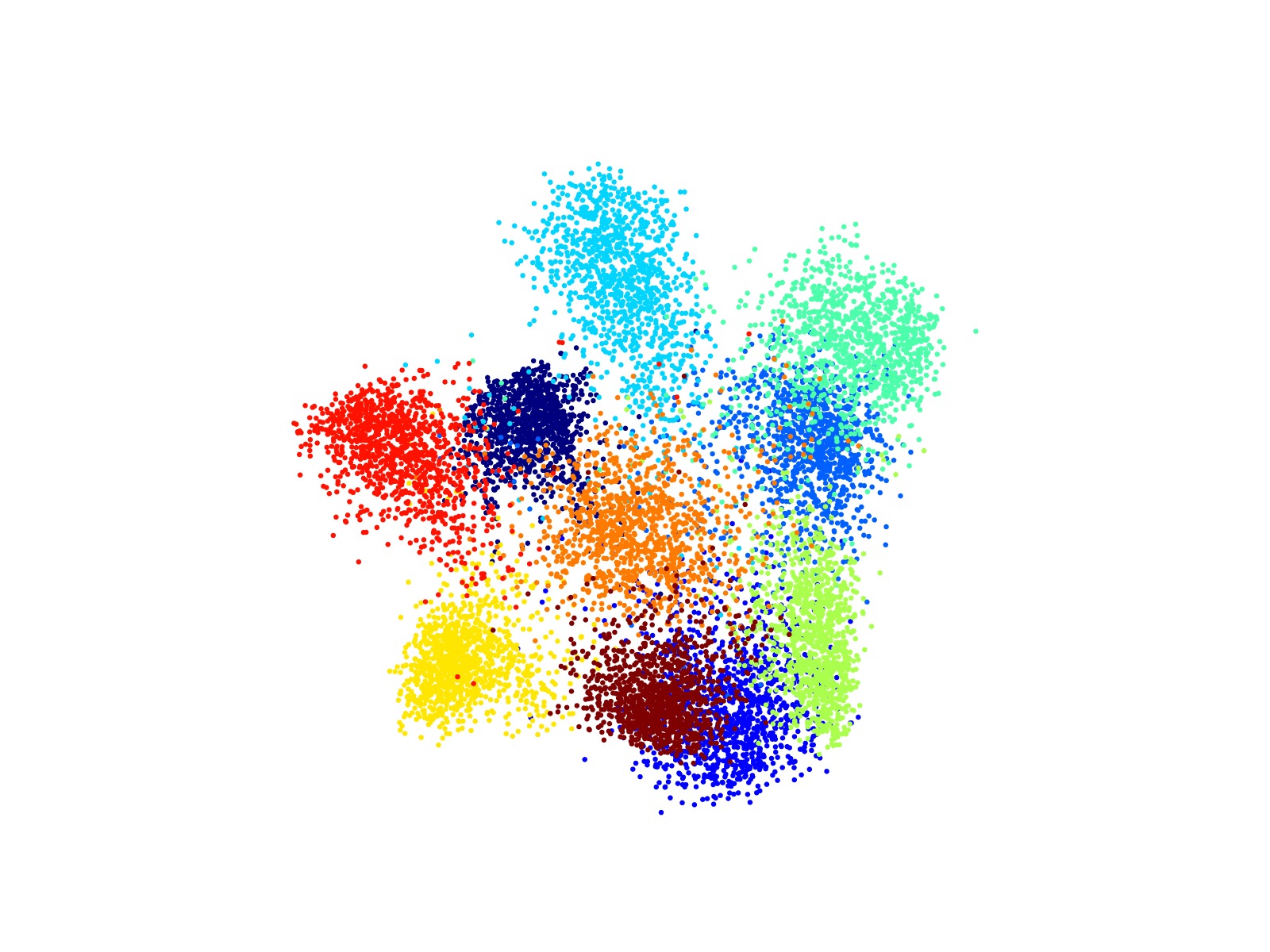}
			\caption{Final stage on \textit{USPS}}
		\end{subfigure}
		\caption{Embedding features in different learning stages on \textit{MNIST-full}, \textit{MNIST-test}, and \textit{USPS} datasets. Three stages including Initial stage, Intermediate stage before adding clustering loss, and Final stage are shown for all datasets.}
		\label{fig:6}
	\end{figure*}

\begin{figure*}[!t]
		\begin{subfigure}{.35\textwidth}
			\centering
			\includegraphics[trim=30mm 20mm 30mm 0mm, clip,  width=1\linewidth]{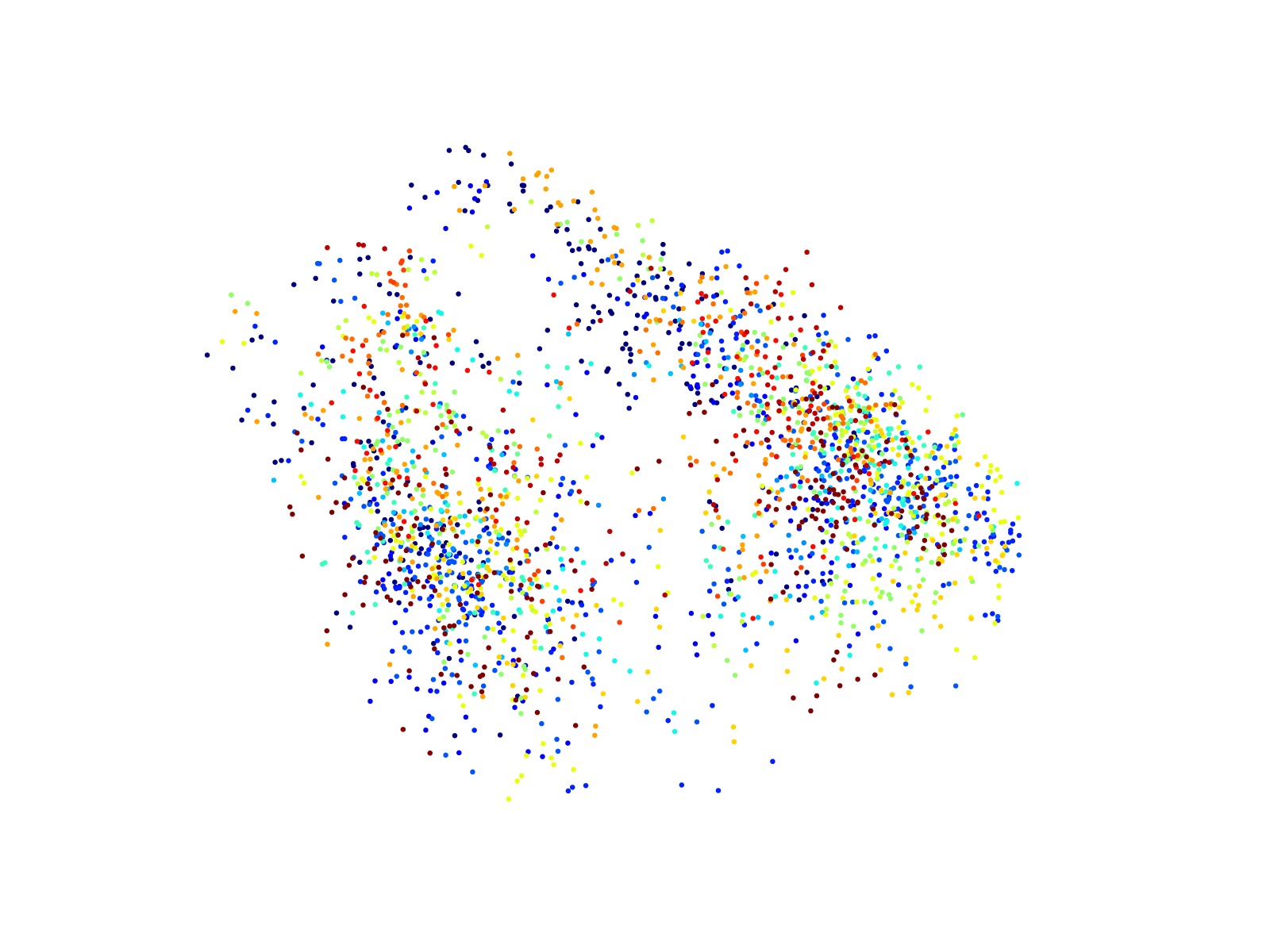}
			\caption{Initial stage on \textit{FRGC}}
		\end{subfigure}%
		\begin{subfigure}{.35\textwidth}
			\centering
			\includegraphics[trim=30mm 20mm 30mm 0mm, clip,  width=1\linewidth]{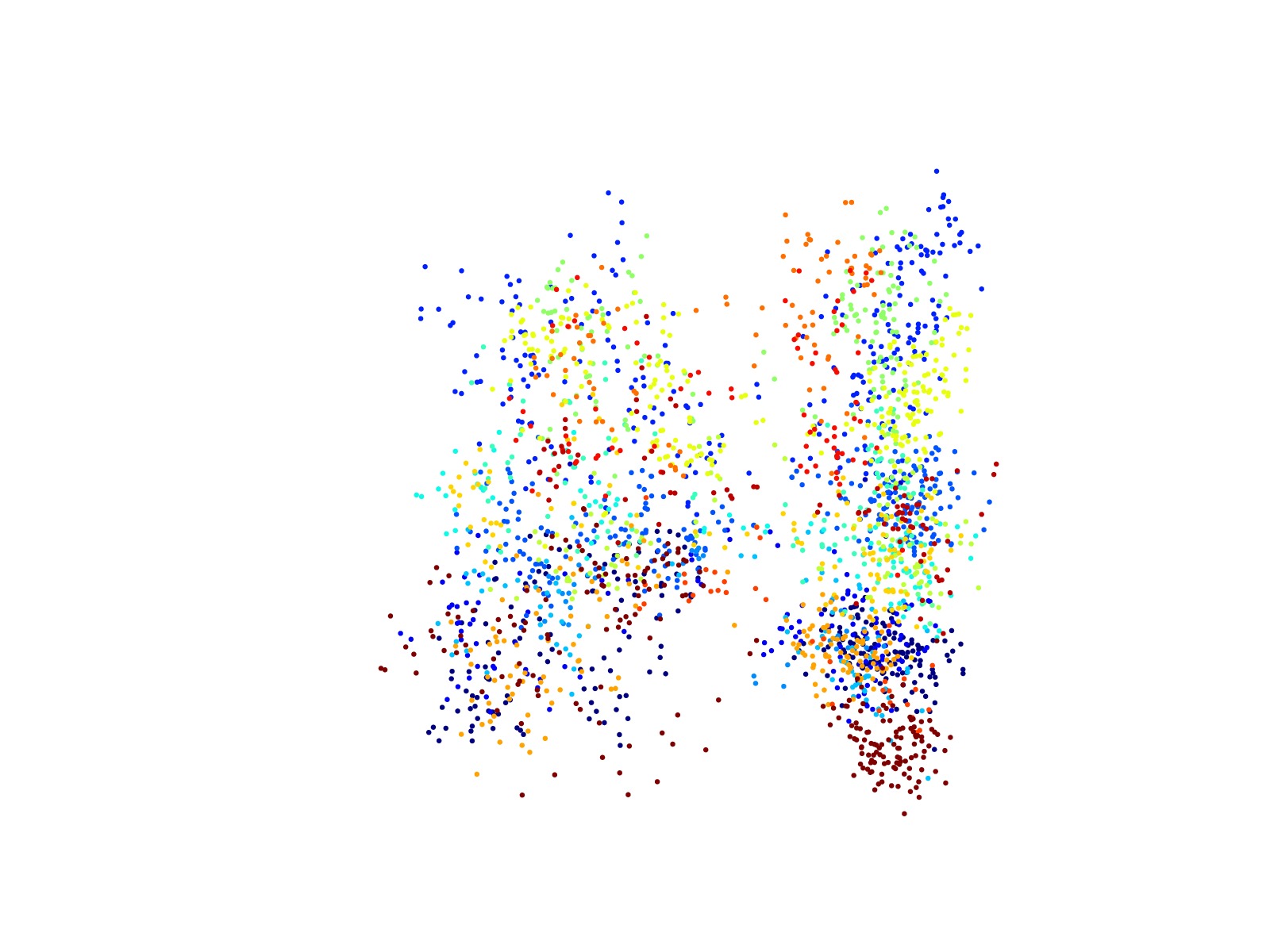}
			\caption{Intermediate stage on \textit{FRGC}}
		\end{subfigure}
		\begin{subfigure}{.35\textwidth}
			\centering
			\includegraphics[trim=30mm 20mm 30mm 0mm, clip,  width=1\linewidth]{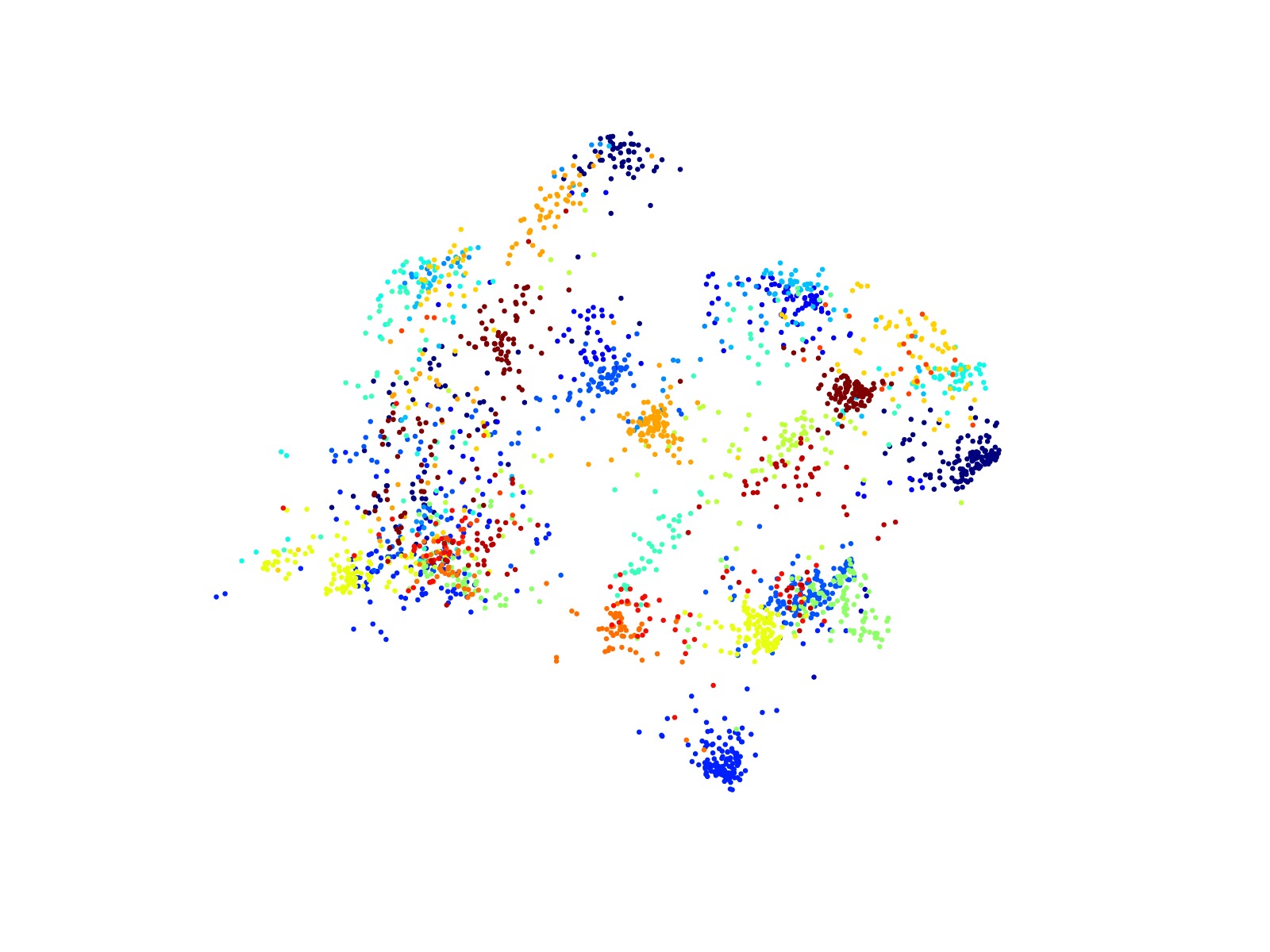}
			\caption{Final stage on \textit{FRGC}}
		\end{subfigure}
		\begin{subfigure}{.35\textwidth}
			\centering
			\includegraphics[trim=30mm 20mm 30mm 0mm, clip,  width=1\linewidth]{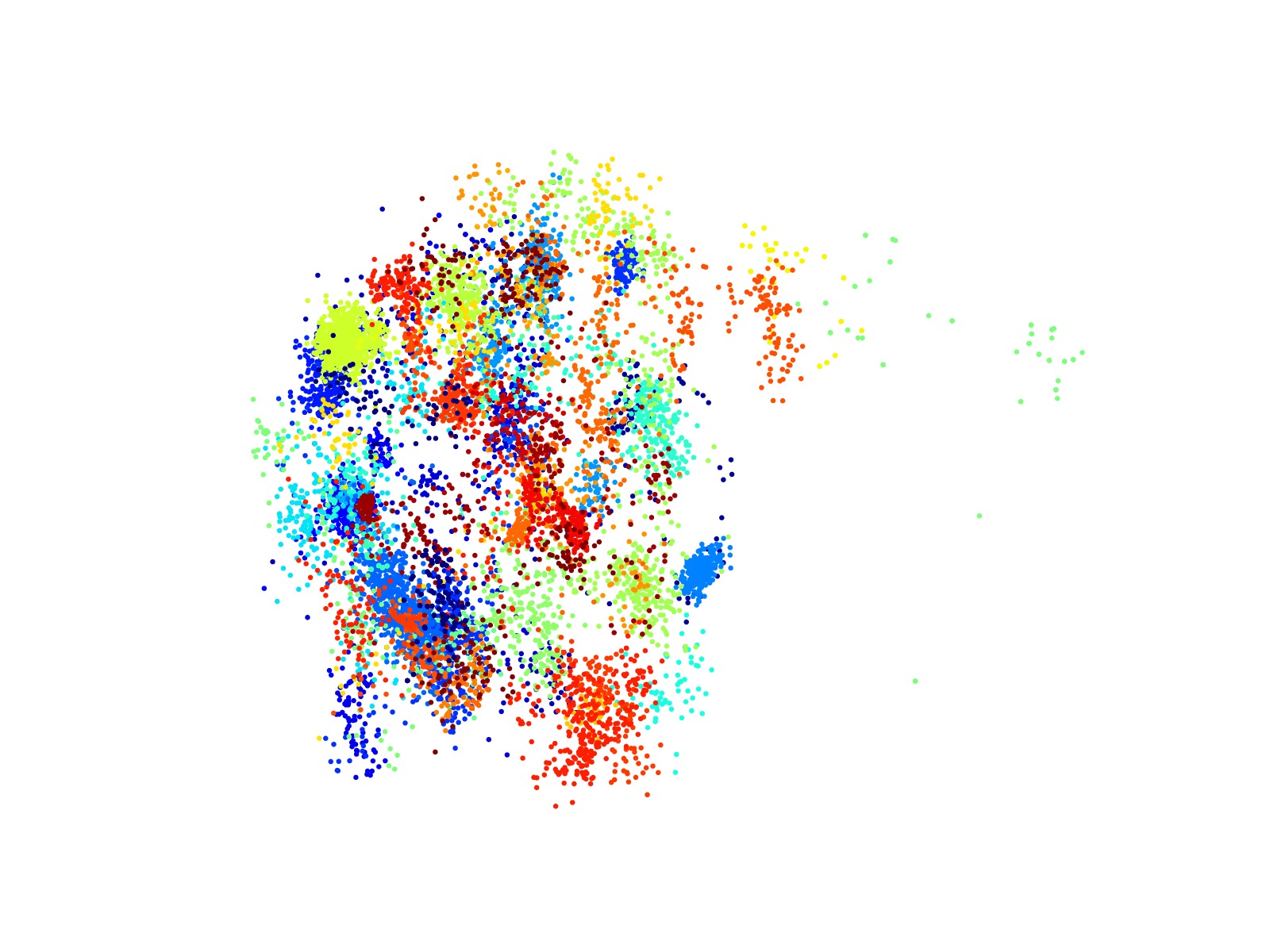}
			\caption{Initial stage on \textit{YTF}}
		\end{subfigure}%
		\begin{subfigure}{.35\textwidth}
			\centering
			\includegraphics[trim=30mm 20mm 30mm 0mm, clip,  width=1\linewidth]{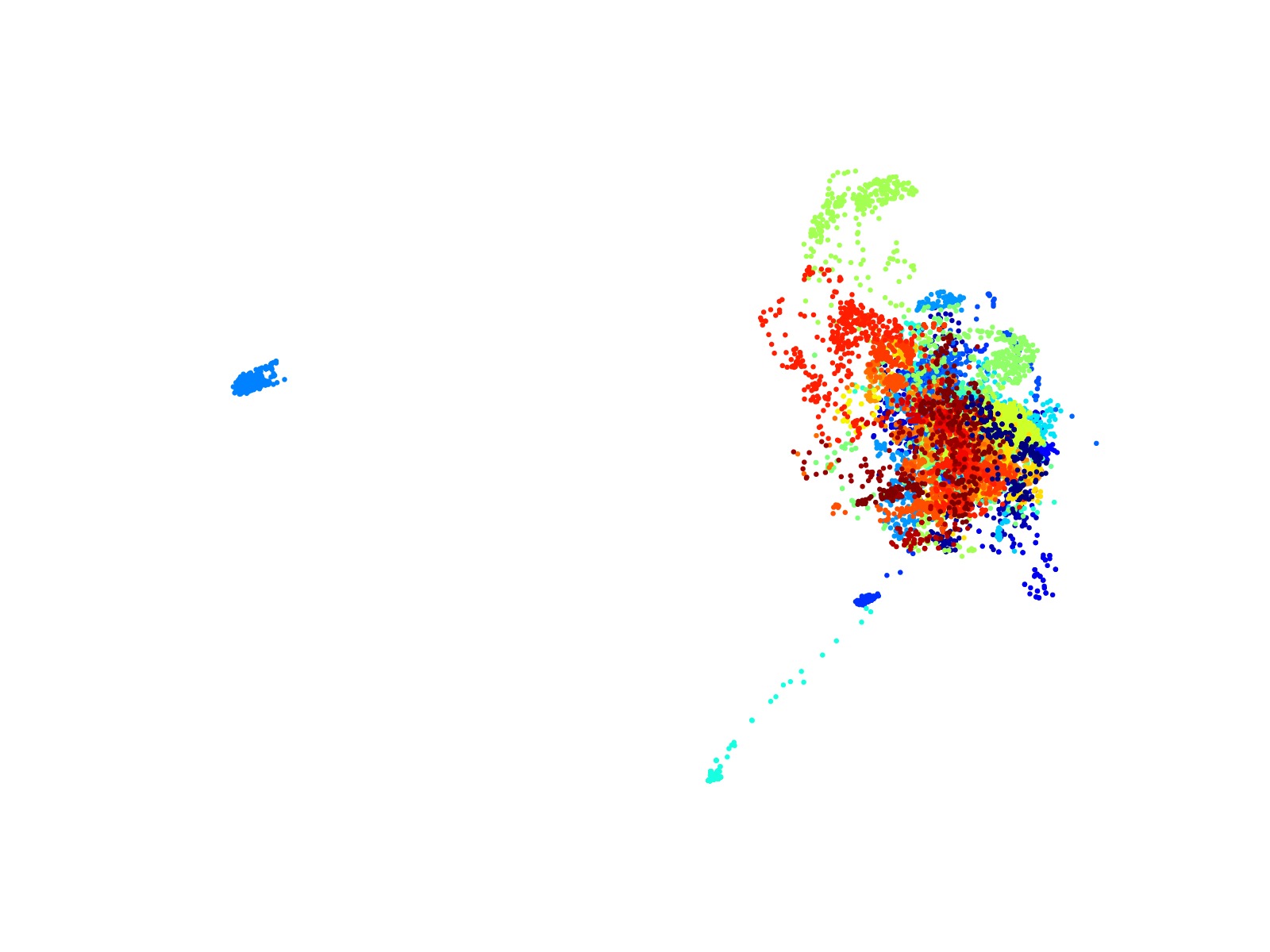}
			\caption{Intermediate stage on \textit{YTF}}
		\end{subfigure}
		\begin{subfigure}{.35\textwidth}
			\centering
			\includegraphics[trim=30mm 20mm 30mm 0mm, clip,  width=1\linewidth]{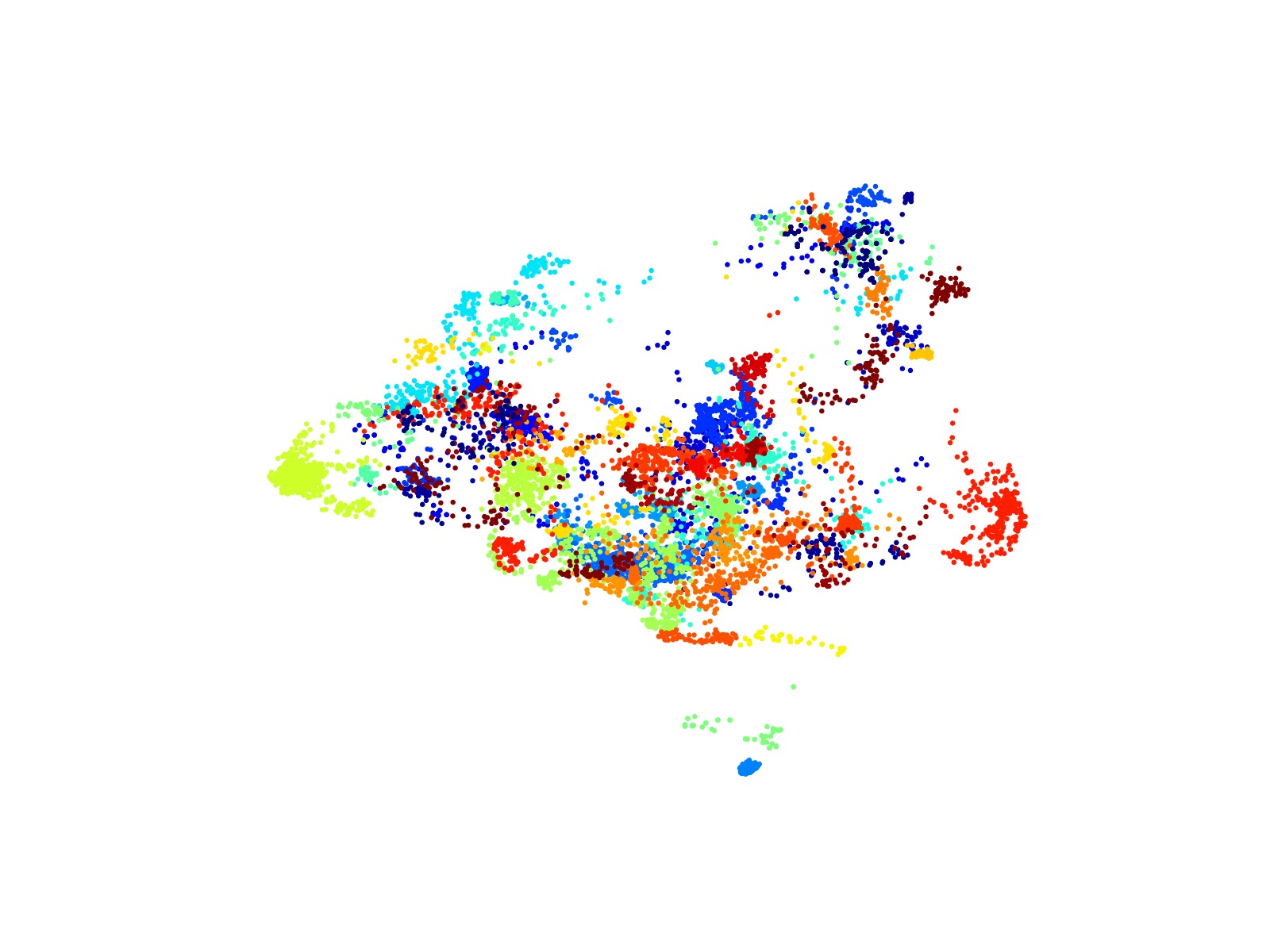}
			\caption{Final stage on \textit{YTF}}
		\end{subfigure}
		\begin{subfigure}{.35\textwidth}
			\centering
			\includegraphics[trim=30mm 20mm 30mm 0mm, clip,  width=1\linewidth]{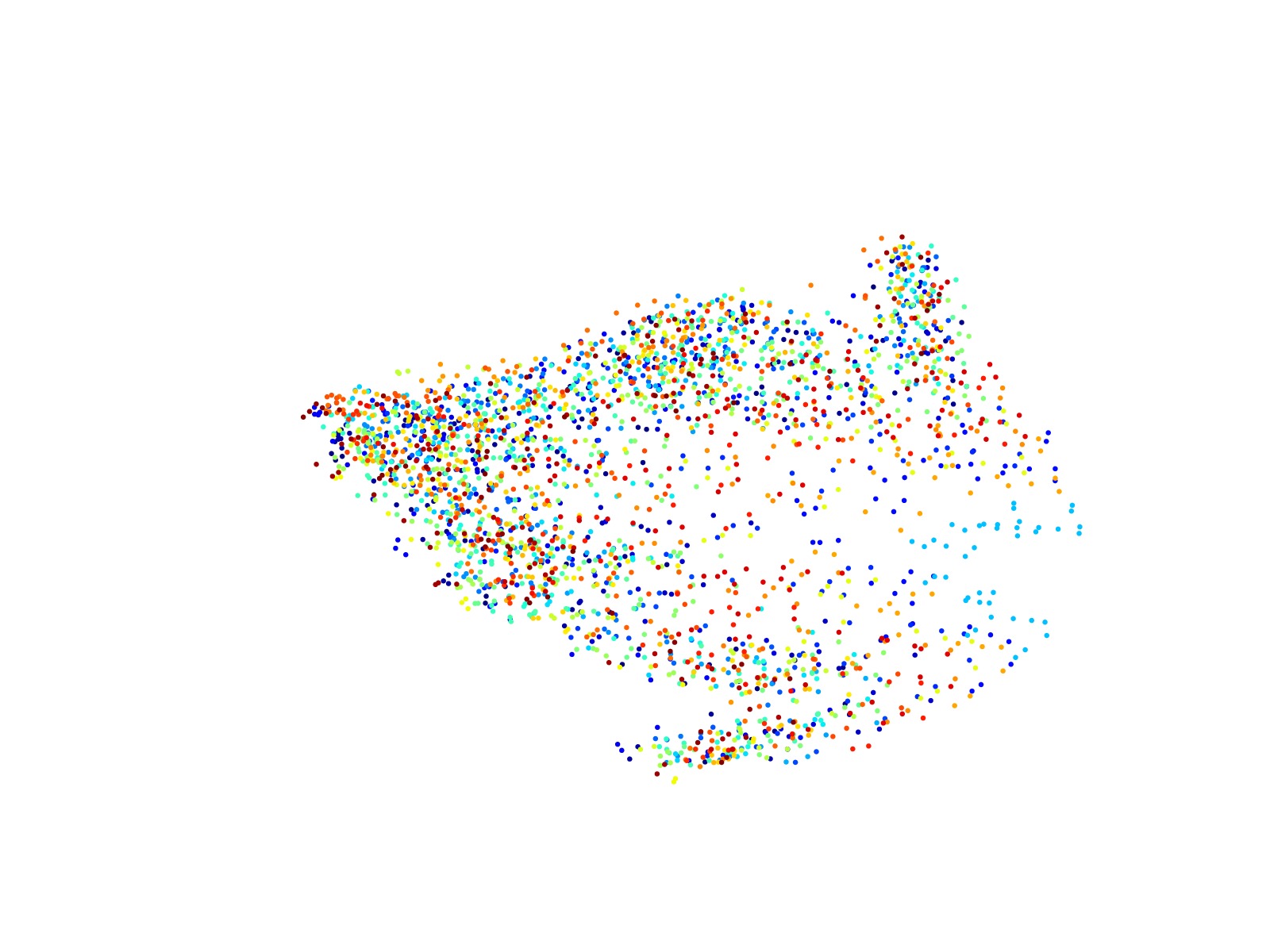}
			\caption{Initial stage on \textit{CMU-PIE}}
		\end{subfigure}%
		\begin{subfigure}{.35\textwidth}
			\centering
			\includegraphics[trim=30mm 20mm 30mm 0mm, clip,  width=1\linewidth]{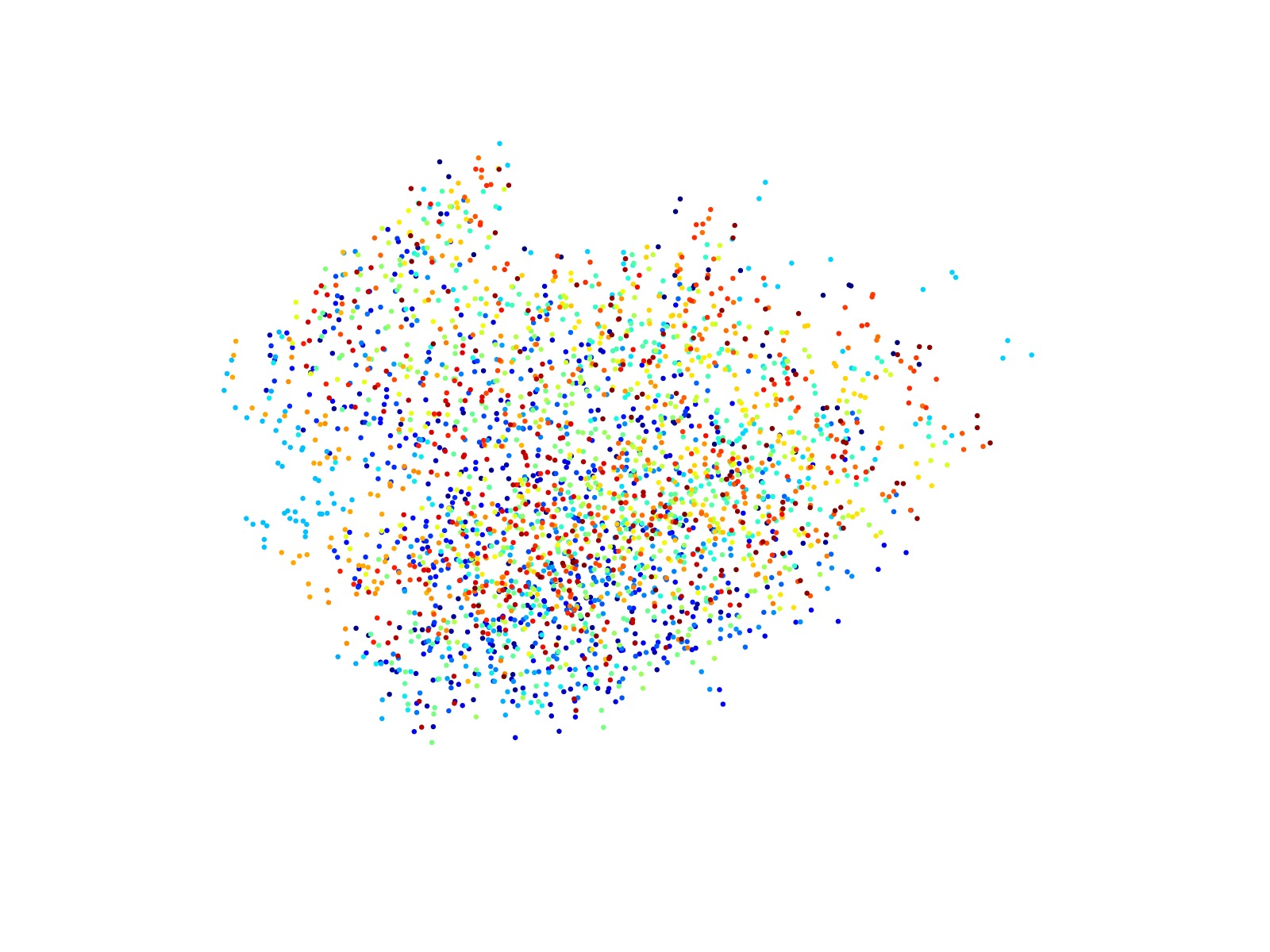}
			\caption{Intermediate stage on \textit{CMU-PIE}}
		\end{subfigure}
		\begin{subfigure}{.35\textwidth}
			\centering
			\includegraphics[trim=30mm 20mm 30mm 0mm, clip,  width=1\linewidth]{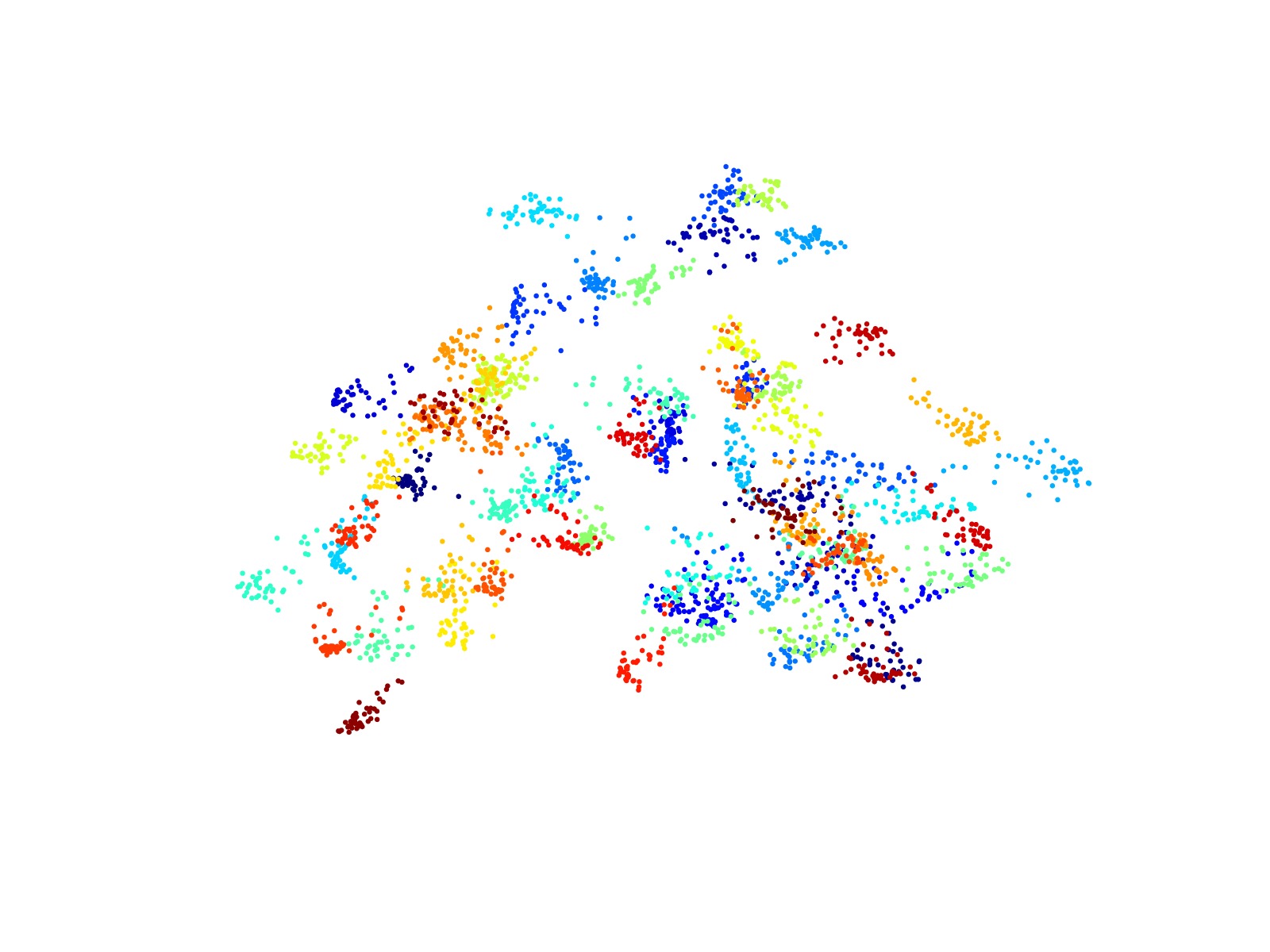}
			\caption{Final stage on \textit{CMU-PIE}}
		\end{subfigure}
		\caption{Embedding features in different learning stages on \textit{FRGC}, \textit{YTF} and \textit{CMU-PIE} datasets. Three stages including Initial stage, Intermediate stage before adding clustering loss, and Final stage are shown for all datasets.}
		\label{fig:7}
	\end{figure*}

{\balance	
	\section{Architecture of Convolutional Autoencoder Networks}
	In this paper, we have two convolutional layers plus one fully connected layer in both encoder and decoder pathways for all datasets. In order to have same size outputs for corresponding convolutional layers in the decoder and encoder,  which is necessary for calculating the reconstruction loss functions, the kernel size, stride and padding (crop in decoder) are varied in different datasets. Moreover, the number of fully connected features (outputs) is chosen equal to the number of clusters for each dataset. Table \ref{table:6} represents the detailed architecture of convolutional autoencoder networks for each dataset.   
	
	
	\section{Visualization of learned embedding subspace}
	In this section, we visualize the learned embedding subspace (top encoder layer) in different stages using the first two principle components. The embedding representations are shown in three stages: 1) initial stage, where the parameters are randomly initialized with GlorotUniform; 2) intermediate stage before adding $L_E$, where the parameters are trained only using reconstruction loss functions; 3) final stage, where the parameters are fully trained using both clustering and reconstruction loss functions. Figure \ref{fig:6} illustrates the three stages of embedding features for \textit{MNIST-full}, \textit{MNIST-test}, and \textit{USPS} datasets, and Figure \ref{fig:7} shows the three stages for \textit{FRGC}, \textit{YTF}, and \textit{CMU-PIE} datasets.
}

\begin{table*}[bp]
	\vspace{-0.0cm}
	\small
	\setlength{\tabcolsep}{3pt}
	\renewcommand{\arraystretch}{0.9}
	\centering
	\begin{tabular}{l|cccc|cccc|c}
		\hline
		Dataset&
		\multicolumn{4}{c|}{Conv1} &
		\multicolumn{4}{c|}{Conv2} &
		Fully \\
		&\# feature maps & kernel size & stride & padding/crop & \# feature maps & kernel size & stride & padding/crop & \# features\\
		\hline
		\textit{MNIST-full}&50&4$\times$4&2&0&50&5$\times$5&2&2&10\\
		\textit{MNIST-test}&50&4$\times$4&2&0&50&5$\times$5&2&2&10\\
		\textit{USPS}&50&4$\times$4&2&0&50&5$\times$5&2&2&10\\
		\textit{FRGC}&50&4$\times$4&2&2&50&5$\times$5&2&2&20\\
		\textit{YTF}&50&5$\times$5&2&2&50&4$\times$4&2&0&41\\
		\textit{CMU-PIE}&50&4$\times$4&2&2&50&5$\times$5&2&2&68\\
		\hline
	\end{tabular}
	\vspace{-0.1cm}
	\caption{Architecture of deep convolutional autoencoder networks. Conv1, Conv2 and Fully represent the specifications of the first and second convolutional layers in encoder and decoder pathways and  the  stacked fully connected layer. }
	\label{table:6}
	\vspace{-1.7cm}
\end{table*}

	
\end{appendices}

\end{document}